\documentclass{midl} 

\usepackage{mwe} 
\usepackage{multirow}
\usepackage{graphicx} 
\usepackage{mathtools}

\jmlrvolume{-- Under Review}
\jmlryear{2026}
\jmlrworkshop{Full Paper -- MIDL 2026 submission}
\editors{Under Review for MIDL 2026}

\title[Spectral Failure Modes]{Revealing Hidden Failure Modes in Chest X-ray Classification via Spectral Domain Analysis}
\midlauthor{\Name{Samuel Halimi}
\orcid{1111-2222-3333-4444} \Email{samuel@azmed.co}\\
\Name{Loic Themyr} \Email{loic@azmed.co}\\
\Name{Arnaud Abreu} \Email{arnaud@azmed.co}\\
\addr 10 rue d'Uzès 75002 Paris
}

\begin{document}

\maketitle

\begin{abstract}
Deep learning models for chest X-ray anomaly detection remain vulnerable to subtle distributional shifts (e.g., acquisition technique, patient-related factors, and preprocessing).
Traditional error analysis often relies on semantic metadata or model embeddings, which can mask low-level signal variations that degrade performance.
In this work, we propose a data-centric framework for automated failure mode discovery using spectral analysis.
We project images into the frequency domain and extract a compact profile summarizing the distribution of signal energy across frequency bands.
By performing unsupervised clustering on these spectral profiles, we demonstrate that model failures are not randomly distributed, but are strongly concentrated within specific spectral clusters.
This method effectively isolates "blind spots", enabling the prediction of model reliability and the discovery of performance-degrading data slices without requiring ground-truth failure annotations.
\end{abstract}

\begin{keywords}
Chest X-ray, Failure Mode Discovery, Spectral Analysis, Model Robustness, Unsupervised Clustering.
\end{keywords}

\section{Introduction}
Deep learning has achieved remarkable success in the interpretation of chest X-rays (CXRs), where algorithms now frequently match or exceed radiologist performance in controlled settings \cite{katzman2024deep, bettinger2024evaluation}.
However, in the context of clinical deployment, models often fail to generalize to data that deviates statistically from their training distribution, resulting in significant and unanticipated performance degradation \cite{zech2018variable, yu2022external}.
Ensuring the reliability of these systems requires the ability to anticipate and identify failure modes before they impact patient care.

Two main approaches have been explored to isolate underperforming samples from unseen data. Metadata stratification uses sensitive categorical attributes to spot biases in the system \cite{gichoya2023ai}, while model-centric techniques examine the embedding of models, either to spot out-of-distribution samples \cite{hong2024out} or to cluster coherent underperfoming slices in the datasets \cite{olesen2024slicing}.
However, both metadata stratification and latent space analysis operate on high-level semantic abstractions and fail to capture subtle signal-level shifts and irregularities that drive performance degradation.


Therefore, in this work, we propose a \emph{data-centric} framework that prioritizes signal characteristics over semantic features or metadata labels.
Inspired by recent studies in the field of domain adaptation and generalization \cite{wang2020high, xu2021fourier}, we hypothesize that some systematic model failures correlate with specific profiles in the frequency domain.
From the \emph{Fourier transform} of an image, we compute a compact representation that summarizes the distribution of signal energy across frequency bands.
In that representation space, we apply an unsupervised clustering method to partition validation data into spectrally coherent subgroups.
Our experiments demonstrate that these spectral clusters might act as predictors of model reliability.
We show that model failures are not uniformly distributed, but are concentrated within specific spectral clusters and effectively reveal "blind spots" in the model’s generalization capability.
This approach allows for the identification of performance-degrading data slices rooted in image physics, enabling predictive reliability estimation without the need for ground-truth failure annotations.

\section{Related work}

\subsection{Slice Discovery in medical image datasets}
To identify systematic failures, the standard approach for medical applications consists in slicing validation data by categorical attributes \cite{seyyed2020chexclusion, larrazabal2020gender, ahluwalia2023subgroup}.
Although essential for fairness auditing, this approach is entirely based on the availability of tabular data.
It remains blind to "hidden stratification" \cite{oakden2020hidden}, where performance degrades due to signal-level characteristics that are not recorded in clinical logs.

Beyond categorical attributes, Out-of-Distribution (OOD) detection uses feature space distances to spot anomalies \cite{lee2018simple, liu2020energy} and is also widely explored in medical image applications \cite{tardy2019uncertainty, berger2021confidence, roy2022does, araujo2023few}.
Although effective in identifying stark outliers, these methods struggle with subtle variations in image acquisition that degrade performance without triggering distance-based alarms \cite{wiles2021fine}.

To uncover coherent high-error subsets of data, recent unsupervised methods apply clustering algorithms in the latent embedding space of models \cite{eyuboglu2022domino, d2022spotlight, olesen2024slicing}.
However, this approach faces a technical paradox: deep networks are explicitly optimized to be invariant to nuisance variables \cite{achille2018emergence}.
Consequently, the latent space often suppresses the irregularities that cause model failure, rendering these failure modes invisible to latent space clustering strategies.

\subsection{Frequency domain analysis in deep learning}
In this work, we look for failure modes in the frequency domain of the images, to remove metadata supervision and dependence on subjective model embeddings to focus only on the intrinsic irregularities of the signal.
Important studies on image frequencies have been conducted in domain generalization (DG) \cite{huang2021fsdr, zhao2022test} and domain adaptation (DA) \cite{huang2021rda, yang2022source} to increase the robustness of deep learning models with respect to frequency perturbations.
These methods alter either the low frequencies \cite{guo2018low} or the high frequencies \cite{wang2020high} of the images to train models on adversarial examples in a data augmentation scheme.
Although not directly related to our approach, the success of these techniques brings evidence that some failure modes of deep learning models are explained by characteristics of the Fourier spectrum of input images.

Unlike DG and DA approaches that manipulate spectra to train robust models, we utilize spectral analysis as a \emph{post-hoc} diagnostic tool.
This allows us to cluster data based on image physics rather than semantics, exposing signal-driven failures that escape standard monitoring.

\section{Methods}
\subsection{Radially Averaged Power Spectrum (RAPS)}
We analyze signal-level variations by projecting each image into the frequency domain.
For an image $x$, we compute its 2-D discrete Fourier transform $F = \mathcal{F}(x)$ and magnitude spectrum $S = \lvert F \rvert$.
To obtain a 1-D orientation-invariant descriptor $P$, we compute the Radially Averaged Power Spectrum (RAPS) of the image following the extraction protocol by \citet{torralba2003statistics}.
The computation of the RAPS is illustrated in \figureref{fig:raps}.

\noindent For a given discrete radius $r \in \mathbb{N}$, let $\Omega(r)$ be the set of frequency coordinates such as:

\begin{equation}
    \Omega(r) = \{(u, v) \in \mathbb{Z}^2 \mid \sqrt{u^2 + v^2} = r\}
\end{equation}

\noindent Then, for a given sampling of $N$ discrete radii $[r_1,\ldots,r_N]$, the $n$-th component $P(n)$, of the RAPS $P$, is computed as the average of the spectral magnitude $S$ over $\Omega(r_n)$:

\begin{equation}
    P(n) = \frac{1}{\lvert \Omega(r_n) \rvert} \sum_{(u, v)\in\Omega(r_n)}S(u, v)
\end{equation}

\noindent We employ the RAPS profile to compress images into compact, rotation-invariant vectors, ensuring that downstream clustering tasks focus on intrinsic statistics rather than superficial variations in object pose.

\begin{figure}[htbp]
\floatconts
  {fig:raps}
  {\caption{Computation of the RAPS of an image. We highlight the contribution of three frequency bands (red, green, blue) in the spectrum to the RAPS representation (dashed lines with corresponding colors).}}
  {\includegraphics[width=1.\linewidth]{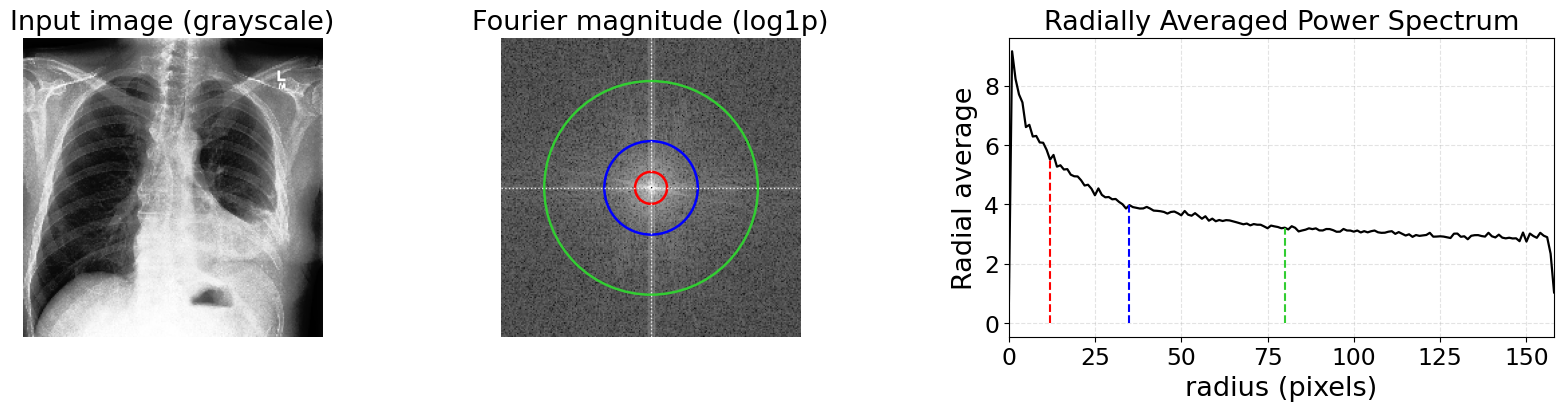}}
\end{figure}

To ensure comparability across images, all Fourier spectra must lie on the same frequency grid.
We therefore restrict the analysis to images of similar native resolution and enforce identical spatial dimensions using minimal spatial cropping.
This avoids spectral distortions: resizing modifies high-frequency content through interpolation, while padding artificially increases low-frequency energy.
A minimal uniform crop preserves the relevant spectral structure and maintains the physical interpretation of Fourier frequencies.

\subsection{Similarity metric}
For the subsequent clustering phase, we advocate the use of the correlation distance as a similarity metric between RAPS descriptors.
Given the RAPS of two different unseen images, $P_i$ and $P_j$, their correlation distance is given by:

\begin{equation}
    d_{\text{corr}}(i,j)
    = 1 - \mathrm{corr}(P_i, P_j)
\end{equation}

\noindent Where the correlation between $P_i$ and $P_j$ is computed across the $N$ sampled radii as:

\begin{equation}
    \mathrm{corr}(P_i, P_j)
    = \frac{\sum_{n=1}^N \big(P_i(n) - \bar P_i\big)\big(P_j(n) - \bar P_j\big)}
    {\sqrt{\sum_{n=1}^N \big(P_i(n) - \bar P_i\big)^2}
           \sqrt{\sum_{n=1}^N \big(P_j(n) - \bar P_j\big)^2}}
\end{equation}

\noindent Unlike the Euclidean distance, which is sensitive to absolute magnitude, the correlation distance is invariant to additive and multiplicative shifts in signal intensity.
This ensures that vectors are grouped based on their intrinsic spectral morphology rather than physically irrelevant variations in global energy or sensor gain.
Consequently, the clustering process effectively isolates stable structural features, such as relative sharpness (shape of the high-frequency tail), sensor or reconstruction noise patterns (boosting of high frequencies), as well as characteristic spectral slopes of specific imaging devices, while remaining robust to acquisition-level scaling and offsets.

\subsection{K-medoids clustering}\label{K-medoids}
In a first exploratory phase, we use the standard K-medoids clustering to validate the relevance of RAPS descriptors for performance analysis. As it is not specifically optimized for slice discovery, it allows unbiased exploration of the relationship between the frequency domain and model failures. The possibility of varying the number of clusters is also convenient to confirm an observed tendency across partitions of different granularities.

\subsection{Hierarchical Agglomerative Clustering (HAC)}\label{HAC}
To optimize the slice discovery process, we resort to a hierarchical clustering technique.
In a bottom-up fashion, each radial profile starts as an individual cluster and we iteratively merge the closest pairs of clusters according to the average linkage rule.
To partition the resulting dendrogram we set the $0.5$-quantile of the distribution of pairwise distances as a threshold to stop the merging procedure.
Adopting this strategy gives three key advantages.
First, the partition scale naturally adapts to the inherent variability of the spectral representations.
Second, it circumvents the limitation of pre-specifying an arbitrary cluster count.
Finally, this flexibility effectively optimizes the isolation and discovery of specific underperforming slices.

\subsection{Outliers}
In accordance with standard agglomerative clustering protocols, outliers were defined as observations or small terminal groups that remained isolated until the final stages of the hierarchy.
Specifically, we classified any cluster containing fewer than $10$ samples as an outlier group, as these represented spectrally atypical profiles with linkage distances significantly exceeding the primary fusion thresholds.
To maintain statistical robustness and ensure sufficient sample sizes for downstream analysis, these isolated profiles were aggregated into a single composite outlier category.
\color{black}

\section{Experimental setup}
\subsection{Datasets and preprocessing}

We conducted all experiments using four publicly available chest X-ray datasets commonly employed for disease classification:

\noindent\textbf{CheXpert} \cite{irvin2019chexpert}. A large-scale dataset from Stanford containing frontal and lateral views, labeled for 14 thoracic findings using report-derived NLP. \textbf{MIMIC-CXR} \cite{johnson2019mimic}. A de-identified dataset from Beth Israel Deaconess Medical Center with paired radiology reports and substantial acquisition and population diversity. \textbf{PadChest} \cite{bustos2020padchest}. A Spanish dataset with multi-view radiographs and detailed labels covering findings, diagnoses, and anatomical locations. \textbf{NIH ChestX-ray14} \cite{wang2017chestx}. A frontal-view dataset labeled for 14 conditions via report mining, notable for heterogeneous acquisition and label noise, making it a common robustness benchmark.

To ensure comparability of Fourier transforms and radial spectral profiles across datasets, we standardized image dimensions while preserving spectral geometry. We selected 264×224 as the target resolution, as it was the most common size across sources. Images were resized isotropically by fixing the height to 224 pixels while preserving aspect ratio. In the Fourier domain, isotropic resizing corresponds to a uniform rescaling of frequencies and avoids the geometric distortions that anisotropic resizing would introduce. This is particularly important for radial averaging, which assumes isotropic spectral structure: anisotropic scaling would induce elliptical distortions and bias the radial profile. Although interpolation during resizing may slightly attenuate high-frequency components, the procedure was applied uniformly across samples, preserving the relative spectral structure underlying our analysis. Minimal center-cropping was then performed to obtain identical final dimensions, and samples with incompatible aspect ratios were excluded to avoid excessive rescaling or truncation.
\color{black}

We focus on three clinically common findings: Atelectasis, Consolidation, and Pleural Effusion. We trained our models in a mono-pathology setting. For each pathology, we constructed separate training and validation splits. The training sets are exclusively based on CheXpert images. Table \ref{tab:dataset} details the datasets sizes. The test sets are built from each dataset, so MIMIC-CXR, PadChest, and NIH are unseen-domain in our evaluation. This enables the analysis of cross-dataset generalization under distributional shift.

For experiments that require large-scale validation, we derived a unified \emph{Global} test set by pooling the test subsets from every datasets.
\color{black}

\begin{table}[htbp]
\floatconts
  {tab:dataset}%
  {\caption{The size of each mono-pathology dataset used in the study is detailed, broken down by the number of images in the training set, the seen-domain test set, and the unseen-seen test set. Note that all datasets are balanced, containing an equal number of positive and negative examples.}}%
  {\scalebox{0.9}{\begin{tabular}{llccc}
\hline
\textbf{Dataset} & \textbf{Type} & \textbf{Atelectasis} & \textbf{Consolidation} & \textbf{Pleural Effusion} \\
\hline
\textbf{CheXpert}                     & Train       & 39K      & 39K      & 39K \\
\hline
\textbf{CheXpert}                     & Test   & 256      & 256      & 256 \\
\hline
\textbf{MIMIC}                        & Test & 476      & 400      & 520 \\
\hline
\textbf{NIH}                          & Test & 1308     & 642      & 1386 \\
\hline
\textbf{Padchest}                     & Test & 400      & 400      & 816 \\
\hline
\textbf{Global}                  & Test & 2440      & 1698      & 2978 \\
\hline
\color{black}
\end{tabular}}}
\end{table}

\subsection{Models, training and performance assessment}
For all experiments, we trained a DenseNet-121 \cite{huang2018denselyconnectedconvolutionalnetworks} classifier using binary cross-entropy loss and the Adam \cite{kingma2017adammethodstochasticoptimization} optimizer (learning rate $1\times 10^{-4}$, weight decay $1\times 10^{-3}$).
A Reduce-on-Plateau scheduler (minimum learning rate $1\times 10^{-5}$) and a batch size of 16 were used.
Models were trained for up to 30 epochs with early stopping based on validation performance.
All training was performed on an NVIDIA GeForce RTX 3080 GPU.

We report the Area Under the Receiver Operating Characteristic (AUROC) to evaluate the discriminative performance of the models.
This metric was chosen for its invariance to decision thresholds and its robustness to the fluctuating prevalence in unsupervised clustering assignments.

\section{Experiments and results}
\subsection{Relevance of RAPS descriptors for performance analysis}\label{relevance}

Given a K-medoids partition of the RAPS descriptors (Section \ref{K-medoids}), we use the standard deviation of the AUROC across the $K$ clusters to measure how effectively the partition stratifies performance.
To isolate the effect of spectral coherence from artifacts of cluster size or class distribution, we implement two stochastic baselines: a \emph{fully random} (FR) assignment and a \emph{pseudo-random} (PR) control that preserves the specific cluster size and label statistics of the K-medoids solution.
To study the impact of granularity, we repeat the above experiment for values of $K$ ranging from $2$ to $10$. We then average the results over 50 seeds to reduce variability from K-medoids initialization and baseline stochasticity.

To demonstrate the consistency of our findings across varying model architectures, we replicated the experiments using both ResNet \cite{he2015deepresiduallearningimage} and EfficientNet \cite{tan2020efficientnetrethinkingmodelscaling} architectures; these results are detailed in Appendix \ref{more_models}.
\color{black}

As shown in \figureref{fig:kmedoids2_corr}, the dispersion of AUROC in the K-medoids partitions of RAPS profiles (blue) remains consistently above those of both random baselines, sometimes reaching nearly twice as high. Importantly, the pseudo-random baseline (red) preserves the exact cluster size distribution and class proportions of the K-medoids solution, while discarding any organization based on spectral similarity. This control ensures that the observed variability cannot be attributed merely to differences in cluster size or label imbalance. 

Moreover, all results are averaged over 50 random seeds to further reduce stochastic effects arising from K-medoids initialization and from random cluster composition. The persistent and substantially larger gap between the blue and red curves indicates that the partitions induced by the correlation distance between RAPS profiles capture genuine heterogeneity in model behavior. 

The consistency of this effect across all pathologies (see Appendix \ref{abla_ext}), across different classifier architectures (ResNet, EfficientNet, and DenseNet; Appendix \ref{more_models}), and for all values of $K$ reinforces our confidence that spectral representations derived from Fourier analysis explain a meaningful portion of the observed performance variability.
\color{black}

Because the correlation distance emphasizes shape differences, this gap suggests that differences in spectral composition—such as the relative distribution of low versus high frequencies, decay rates, or local spectral slopes—are systematically associated with differences in model performance.

\begin{figure}[htbp!]
\floatconts
  {fig:kmedoids2_corr}
  {\caption{AUROC variability across different number of clusters ($K$) made by the k-medoids method with the correlation metric vs. clusters made by fully and pseudo random methods. Results are averaged over 50 seeds. Clustering is performed on the Global aggregated test set.}}
  {%
    \centering
    \begin{tabular}{ccc}
      \includegraphics[width=0.3\linewidth]{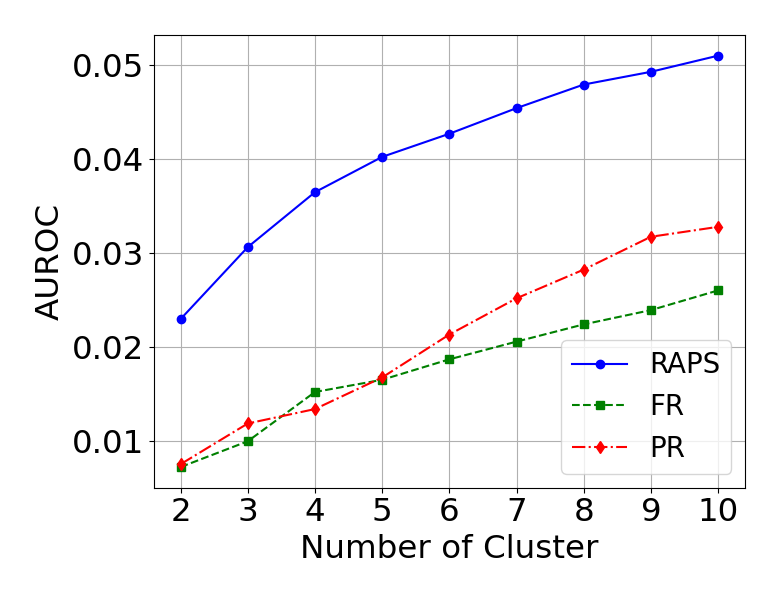} &
      \includegraphics[width=0.3\linewidth]{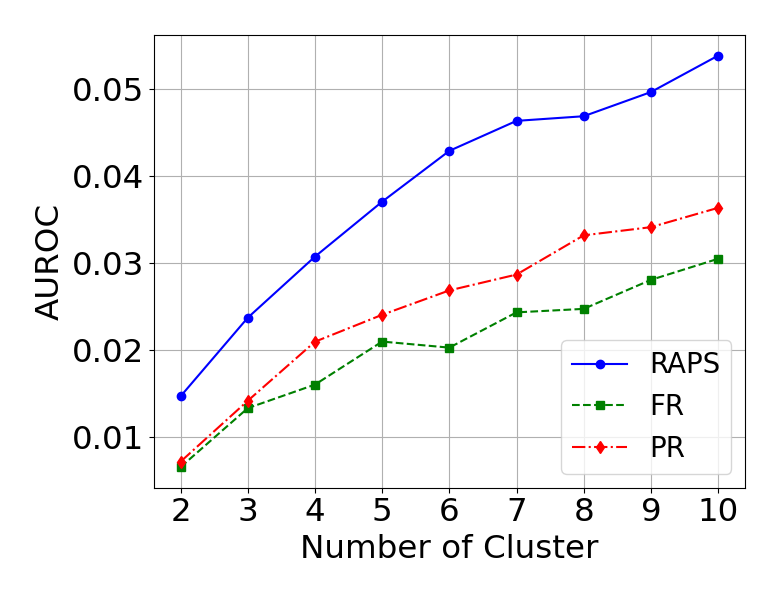} &
      \includegraphics[width=0.3\linewidth]{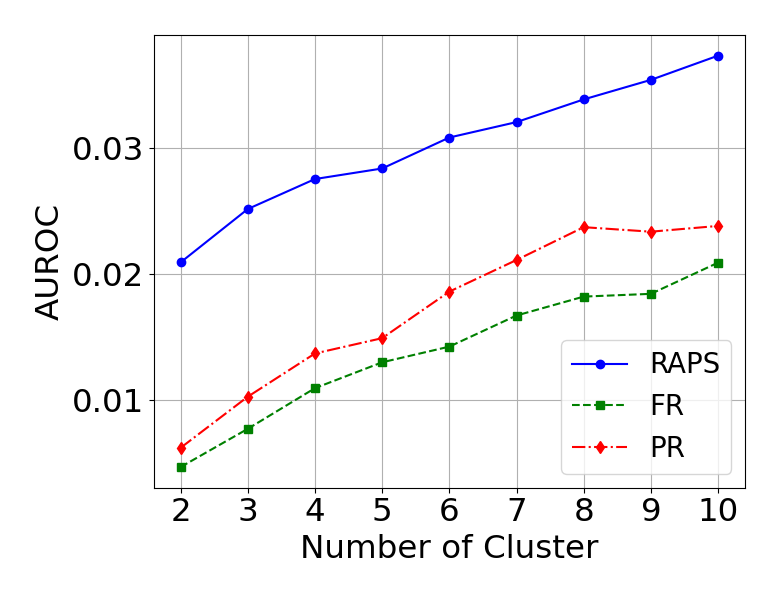} \\
      (a) Atelectasis & (b) Consolidation & (c) Pleural Effusion
    \end{tabular}
  }
\end{figure}

\subsection{Influence of the similarity metric on performance stratification}

To assess how the distance metric influences the K-medoids performance stratification, we compare three metrics on the radial profiles: Euclidean, Spearman, and correlation. The evaluation follows the same protocol as in Section \ref{relevance}, with AUROC variability compared to the FR and PR baselines and averaged over multiple seeds.

Across all values of $K$, the correlation distance produces the largest separation between the clustering curve and both baselines (\figureref{fig:ablation_atelectasis}). The Spearman distance induces greater variability than the Euclidean distance, though still below the correlation distance.

By contrast, the $L_2$ distance shows only limited gains over the baselines, suggesting that amplitude-based similarities contribute little to the stratification of performance compared to differences in the \emph{shape} of the spectral profiles.
This ordering among metrics indicates that the performance variability is more closely related to the \emph{relative distribution} of spectral energy than to its absolute magnitude.
Metrics that are invariant to global scaling and affine transformations, such as the correlation distance and, to a lesser extent, the Spearman distance, capture spectral characteristics that are more informative for distinguishing model behavior.
This supports the choice of the correlation distance in our clustering framework and is consistent with Section \ref{relevance}, where it produces the largest performance disparities across clusters.

These findings suggest that DG and DA strategies based on the frequency domain should not focus exclusively on matching absolute spectral energy distributions, as is often done in methods that impose target spectra or mix frequency components directly. Instead, our results indicate that the \emph{relative} structure of the radial spectrum, such as the balance between low and high frequencies and the rate at which spectral energy decays, plays a more consequential role. Methods that explicitly account for these relative spectral characteristics may therefore offer a more effective direction for adaptation and robustness.

\begin{figure}[htbp]
\floatconts
  {fig:ablation_atelectasis}
  {\caption{Ablation study evaluating the influence of the distance metric used in k-medoids clustering on AUROC variation for the Atelectasis pathology. We compare three metrics: Euclidean ($L_2$) distance, Spearman distance, and the correlation distance employed in our proposed method.}}
  {%
    \centering
    \begin{tabular}{ccc}
      \includegraphics[width=0.3\linewidth]{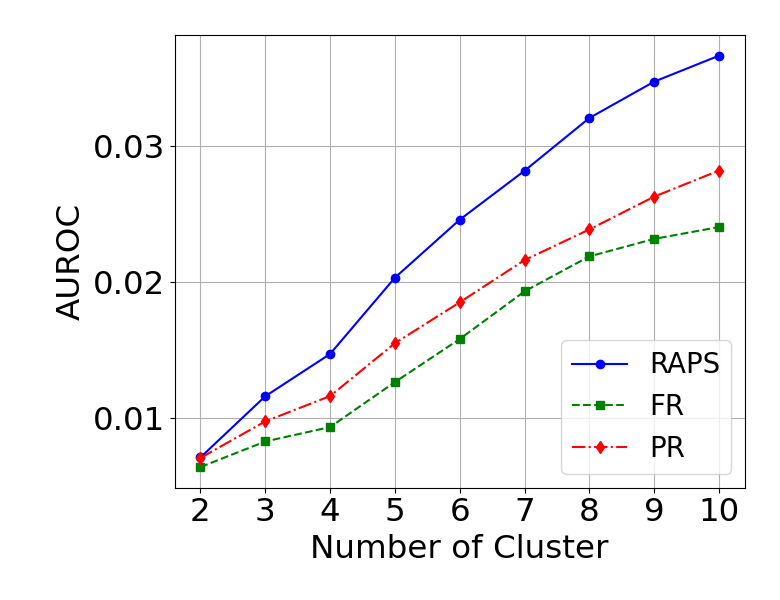} &

      \includegraphics[width=0.3\linewidth]{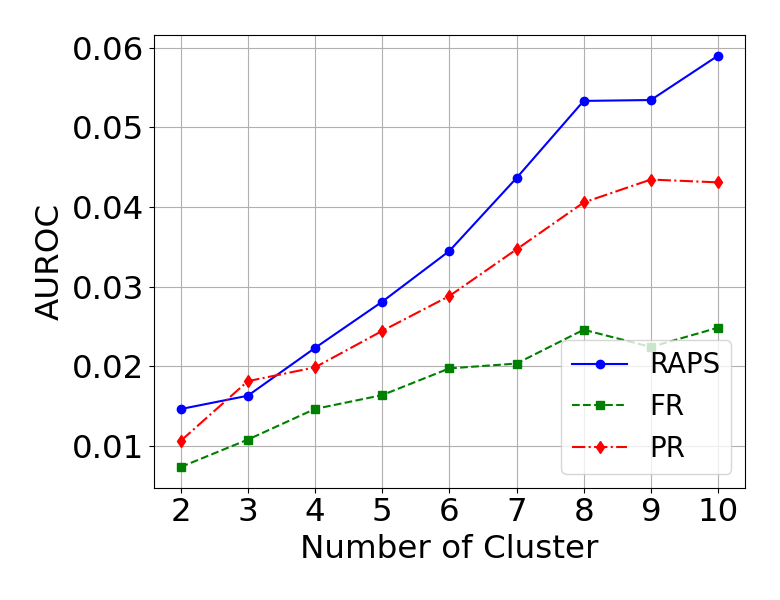} &

      \includegraphics[width=0.3\linewidth]{images/exp2/kmedoids/corr/kmedoids_plot_correlation__ONLYCHEXPERT_CORR_all_seed_std_stats_n50_filtred_train_atelectasis_landscape_fromImagenet.png} \\
      
      (a) $L_2$ & (b) Spearman & (c) Correlation
    \end{tabular}
  }
\end{figure}

\subsection{Slice discovery and outlier detection}

To confirm the robustness of our data-centric framework, we apply HAC in the space of RAPS descriptors to isolate under-performing data slices on unseen domain data.
After clustering, we assess the AUROC on each cluster to pinpoint specific spectral regions where model performance deviates from the norm.

\begin{figure}[htbp]
\floatconts
  {fig:outliers_2}
  {\caption{Comparative analysis of model performance (AUROC) across data clusters (blue) defined by HAC. The figure highlights the distinct model reliability on the identified clusters versus the significantly different performance (red) observed on the HAC-isolated outlier set. Results on the unseen-domain test datasets.}}
  {%
    \centering
    \begin{tabular}{c}
      \includegraphics[width=1.\linewidth]{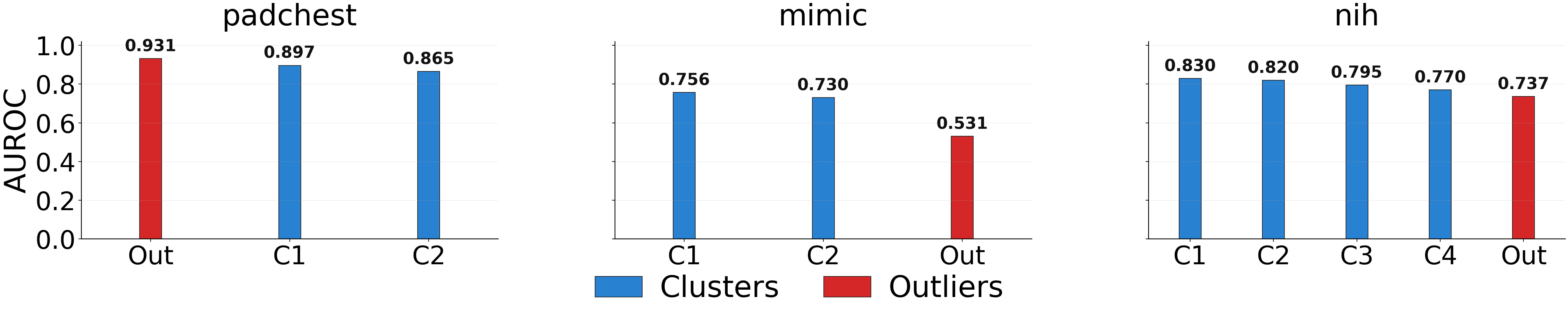} \\
      (a) Atelectasis \\
      \includegraphics[width=1.\linewidth]{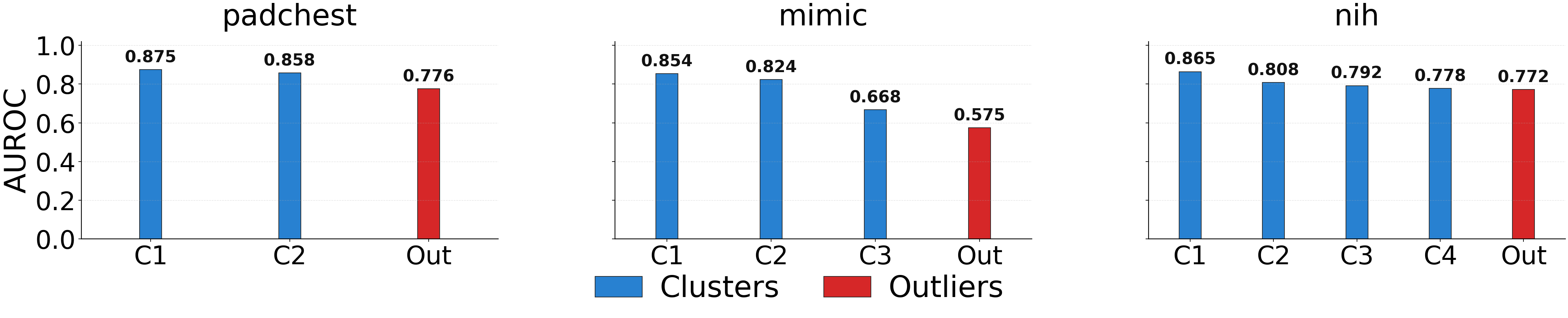} \\
      (b) Consolidation \\
      \includegraphics[width=1.\linewidth]{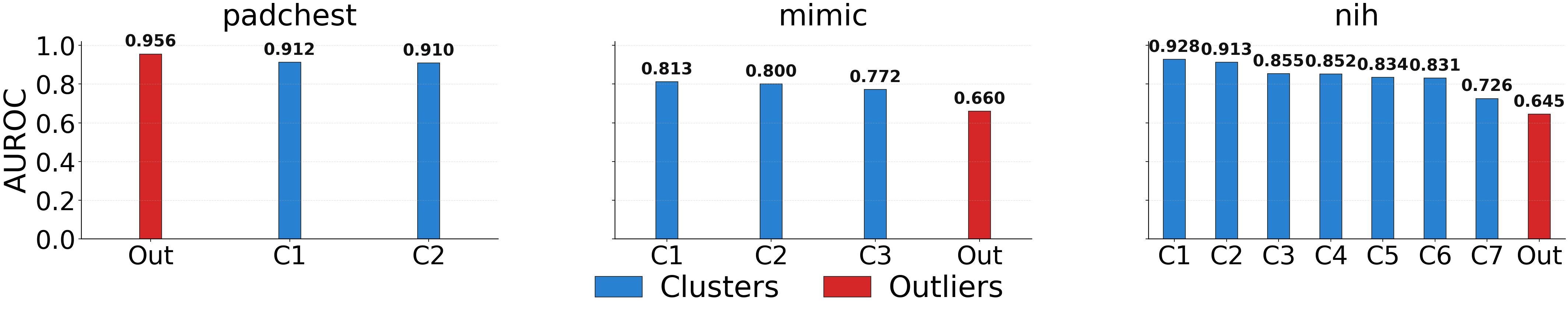} \\
      (c) Pleural Effusion \\
    \end{tabular}
  }
\end{figure}

As depicted in \figureref{fig:outliers_2}, the AUROC across HAC clusters exhibits distinct and variable performance levels for all pathologies and external validation centers.
This finding mirrors the performance stratification observed with the K-medoids clustering approach.
This convergence confirms that the spectral differences encoded through the correlation distance reliably capture meaningful variations in model performance, regardless of the underlying clustering algorithm employed.

Importantly, we demonstrate that across almost all considered datasets and pathologies, the proposed method systematically isolates a cluster exhibiting significantly degraded AUROC (bars on the far right of the different plots).
This consistency underscores the efficacy of applying HAC to RAPS descriptors for slice discovery, as it reliably uncovers latent subgroups where model performance is compromised.

Additionally, a particularly characteristic and instructive pattern emerged from the analysis of the outlier super-group (red), which was formally defined by merging all HAC clusters containing fewer than $10$ samples. The AUROC for this super-group consistently exhibited a marked separation from the performance mean of the main clusters. Specifically, the performance was frequently significantly lower; as observed, for example, in Atelectasis and Pleural Effusion on the MIMIC and NIH datasets, and in Consolidation on Padchest and MIMIC; indicating substantial failure concentration. Conversely, in select cases, the AUROC was noticeably higher (e.g., Atelectasis and Pleural Effusion on Padchest), but critically, it was never comparable to the average cluster performance.

As suggested by the analyses in Appendix \ref{slice_comparison_apendix} and \ref{quali}, these variations in performance are not readily attributable to observable visual or demographic differences alone. \color{black} Therefore, this consistent performance separation provides strong quantitative evidence: slices possessing atypical spectral signatures reliably translate into performance-level outliers within the model.

\begin{figure}[htbp]
\floatconts
  {fig:seen_hac}
  {\caption{Comparative analysis of model performance (AUROC) across data clusters (blue) defined by HAC. The figure highlights the distinct model reliability on the identified clusters versus the significantly different performance (red) observed on the HAC-isolated outlier set. Results on seen-domain test datasets: CheXpert.}}
  {%
    \centering
    \begin{tabular}{ccc}
      \includegraphics[width=0.3\linewidth]{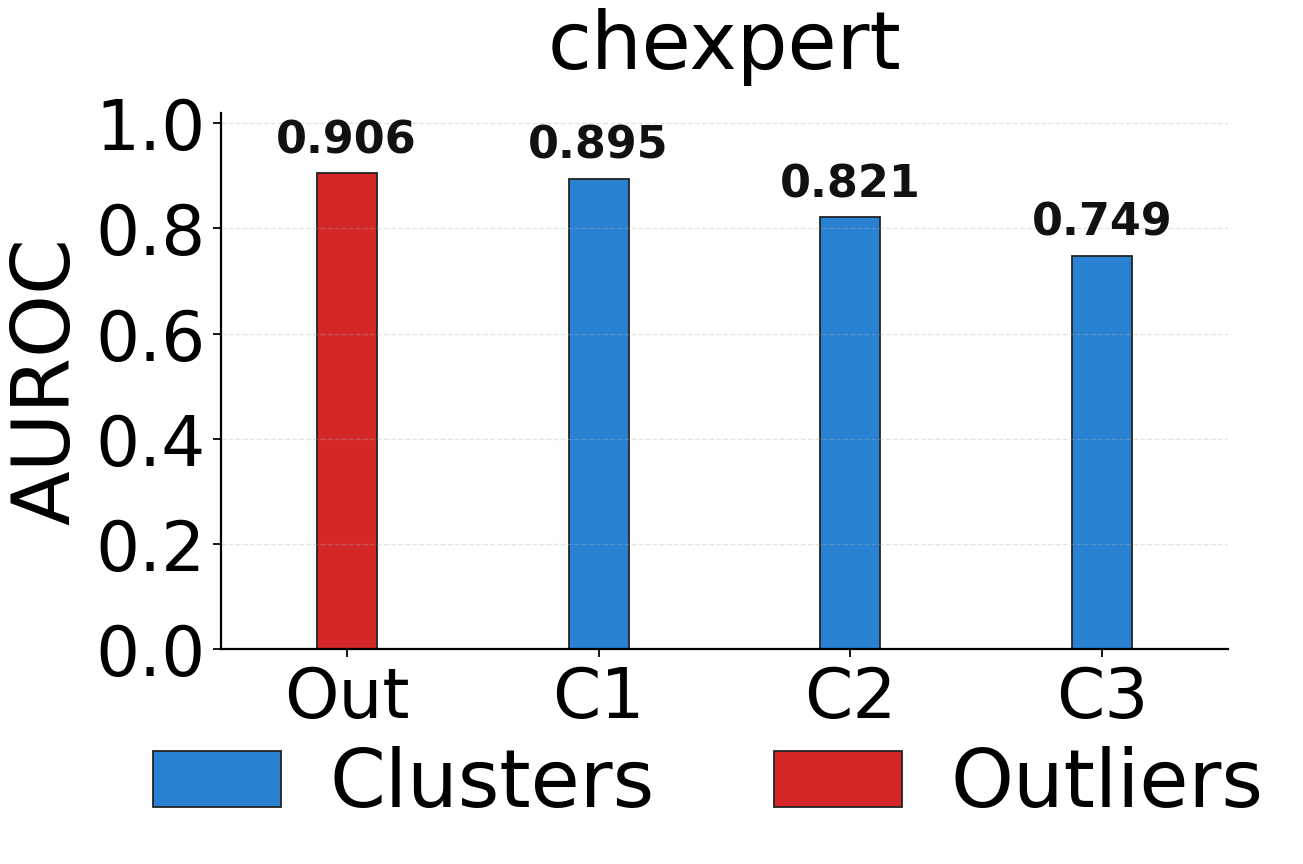} &
      \includegraphics[width=0.3\linewidth]{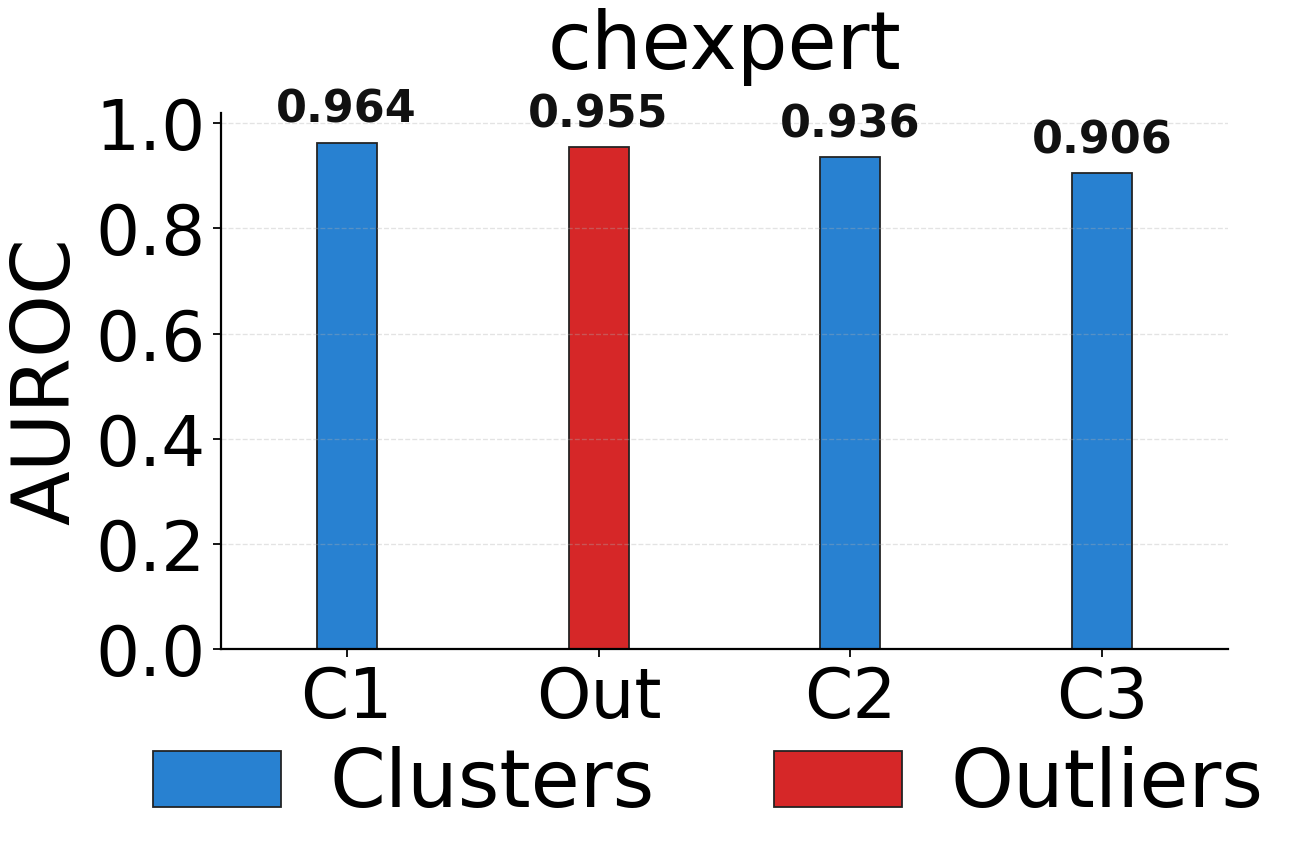} &
      \includegraphics[width=0.3\linewidth]{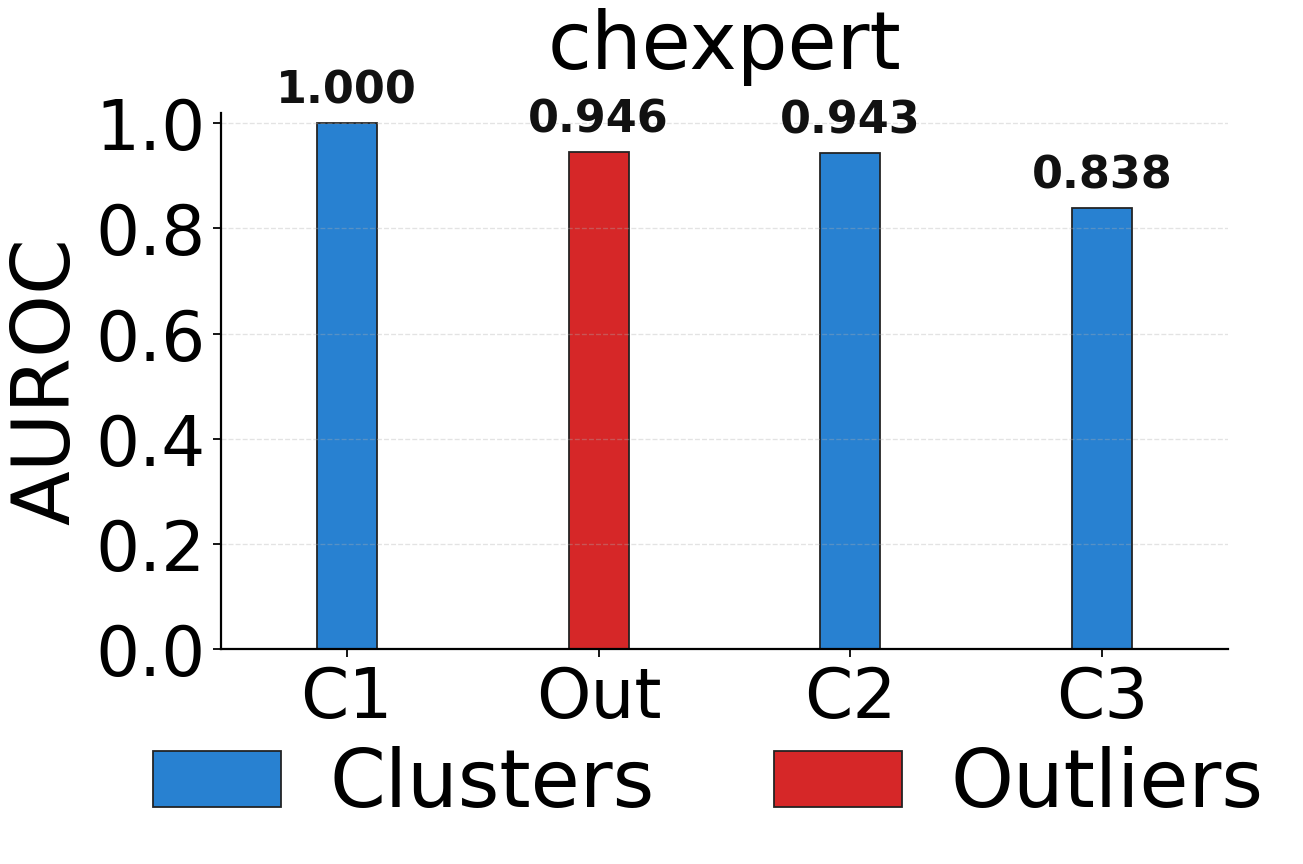} \\
      
      (a) Atelectasis & (b) Consolidation & (c) Pleural Effusion
    \end{tabular}
  }
\end{figure}

We also evaluated the model on seen-domain images with the CheXpert test dataset. As shown in \figureref{fig:seen_hac}, the performance on the outlier super-group is not markedly different from that of the other clusters. This outcome is expected: because the model was trained exclusively on this domain, it learned to handle the full range of RAPS variations present in the training distribution. Nevertheless, we still observe clusters with distinct performance levels, indicating that HAC can partition the domain into meaningful RAPS subsets on which the model performs comparatively better or worse.

\subsection{Comparison with OOD detection baselines}

Beyond internal performance stratification, we further evaluate whether the proposed spectral framework can effectively identify performance-degrading samples in an out-of-distribution (OOD) setting. To this end, we compare RAPS-based outlier detection against established OOD baselines: \cite{lee2018simple} and \cite{tardy2019uncertainty}. Table \ref{tab:consolidated_comparison} presents the AUROC difference ($\Delta$) relative to the global test set for each experiment. Our results show that HAC of RAPS more consistently isolates OOD clusters with performance levels significantly lower than the baseline. This trend is further evidenced by the mean absolute difference, where HAC of RAPS achieves 0.103, compared to 0.077 and 0.053 for the baselines. Importantly, while both SOTA methods are supervised, HAC of RAPS is fully unsupervised, making it independent to the annotation or borderline cases.
\color{black}

\begin{table}[!htbp]
\floatconts
  {tab:consolidated_comparison}
  {\caption{Performance Baselines (Test) and Method Deltas (OOD - Test). The $\Delta$ represents the difference between the AUROC on detected outliers and the global test set baseline.}}
  {%
    \footnotesize 
    \setlength{\tabcolsep}{3.5pt} 
    \begin{tabular}{l | ccc | ccc | ccc | c}
        \hline
        & \multicolumn{3}{c|}{\textbf{Consolidation}} & \multicolumn{3}{c|}{\textbf{Atelectasis}} & \multicolumn{3}{c|}{\textbf{Pleural Effusion}} & \textbf{AVG} \\
        \textbf{Method} & \textbf{Pad.} & \textbf{MIM.} & \textbf{NIH} & \textbf{Pad.} & \textbf{MIM.} & \textbf{NIH} & \textbf{Pad.} & \textbf{MIM.} & \textbf{NIH} &  $|\Delta|$ \\
        \cline{1-10}
        Test (Baseline) & 0.848 & 0.765 & 0.793 & 0.888 & 0.739 & 0.778 & 0.920 & 0.787 & 0.835 & \\
        \hline
        HAC of RAPS ($\Delta$) & \textbf{-.072} & \textbf{-.190} & -.021 & +.043 & \textbf{-.209} & -.041 & +.036 & \textbf{-.127} & \textbf{-.190} & \textbf{0.103} \\
        \cite{lee2018simple} ($\Delta$) & +.023 & +.076 & +.029 & \textbf{-.124} & -.139 & \textbf{+.110} & -.063 & -.052 & -.081 & 0.077 \\
        \cite{tardy2019uncertainty} ($\Delta$) & -.053 & +.033 & \textbf{+.039} & -.087 & -.032 & +.049 & \textbf{-.066} & -.099 & -.071 & 0.053 \\
        \hline
    \end{tabular}
  }
\end{table}
\color{black}

\clearpage

\section{Conclusion}

In this work, we introduced a data-centric framework that uses Fourier-domain analysis to examine performance variability in chest X-ray classifiers. Representing each image through its Radially Averaged Power Spectrum (RAPS) yielded model-agnostic descriptors capturing acquisition-level signal properties. Across experiments, K-medoids and hierarchical clustering consistently revealed meaningful differences in model behavior across groups defined by their spectral profiles, with hierarchical clustering further isolating small sets of images whose atypical frequency patterns were matched by equally atypical performance across evaluation domains.

Ablations over similarity metrics clarified which components of the spectral representation matter most. Correlation distance produced the strongest separation between clusters, indicating that performance differences relate primarily to the relative distribution of spectral energy, its shape and rate of decay, rather than to absolute magnitude. This identifies the spectral characteristics most relevant to model behavior and motivates approaches that explicitly account for such relative frequency patterns.

Together, these findings point to limitations of metadata- or embedding-based slice discovery. Signal-level irregularities rooted in acquisition physics, preprocessing, or frequency composition form predictive patterns that standard monitoring pipelines overlook. Spectral analysis therefore provides a principled and interpretable means of exposing these hidden strata and improving performance auditing. Beyond error analysis, our results suggest that robustness and adaptation strategies may benefit from targeting spectral structure directly, particularly the balance between low and high frequencies, offering a more physically grounded avenue for improving generalization in medical imaging.


\clearpage  
\midlacknowledgments{The authors thank Alexandre Attia, Julien Vidal and Elie Zerbib for supporting this work at AZmed.}

\bibliography{midl-samplebibliography}

\begin{thebibliography}{37}
\providecommand{\natexlab}[1]{#1}
\providecommand{\url}[1]{\texttt{#1}}
\expandafter\ifx\csname urlstyle\endcsname\relax
  \providecommand{\doi}[1]{doi: #1}\else
  \providecommand{\doi}{doi: \begingroup \urlstyle{rm}\Url}\fi

\bibitem[Achille and Soatto(2018)]{achille2018emergence}
Alessandro Achille and Stefano Soatto.
\newblock Emergence of invariance and disentanglement in deep representations.
\newblock \emph{Journal of Machine Learning Research}, 19\penalty0
  (50):\penalty0 1--34, 2018.

\bibitem[Ahluwalia et~al.(2023)Ahluwalia, Abdalla, Sanayei, Seyyed-Kalantari,
  Hussain, Ali, and Fine]{ahluwalia2023subgroup}
Monish Ahluwalia, Mohamed Abdalla, James Sanayei, Laleh Seyyed-Kalantari,
  Mohannad Hussain, Amna Ali, and Benjamin Fine.
\newblock The subgroup imperative: chest radiograph classifier generalization
  gaps in patient, setting, and pathology subgroups.
\newblock \emph{Radiology: Artificial Intelligence}, 5\penalty0 (5):\penalty0
  e220270, 2023.

\bibitem[Araujo et~al.(2023)Araujo, Aresta, Schmidt-Erfurth, and
  Bogunovi{\'c}]{araujo2023few}
Teresa Araujo, Guilherme Aresta, Ursula Schmidt-Erfurth, and Hrvoje
  Bogunovi{\'c}.
\newblock Few-shot out-of-distribution detection for automated screening in
  retinal oct images using deep learning.
\newblock \emph{Scientific Reports}, 13\penalty0 (1):\penalty0 16231, 2023.

\bibitem[Berger et~al.(2021)Berger, Paschali, Glocker, and
  Kamnitsas]{berger2021confidence}
Christoph Berger, Magdalini Paschali, Ben Glocker, and Konstantinos Kamnitsas.
\newblock Confidence-based out-of-distribution detection: a comparative study
  and analysis.
\newblock In \emph{International Workshop on Uncertainty for Safe Utilization
  of Machine Learning in Medical Imaging}, pages 122--132. Springer, 2021.

\bibitem[Bettinger et~al.(2024)Bettinger, Lenczner, Guigui, Rotenberg, Zerbib,
  Attia, Vidal, and Beaumel]{bettinger2024evaluation}
Hubert Bettinger, Gregory Lenczner, Jean Guigui, Luc Rotenberg, Elie Zerbib,
  Alexandre Attia, Julien Vidal, and Pauline Beaumel.
\newblock Evaluation of the performance of an artificial intelligence (ai)
  algorithm in detecting thoracic pathologies on chest radiographs.
\newblock \emph{Diagnostics}, 14\penalty0 (11):\penalty0 1183, 2024.

\bibitem[Bustos et~al.(2020)Bustos, Pertusa, Salinas, and De~La
  Iglesia-Vaya]{bustos2020padchest}
Aurelia Bustos, Antonio Pertusa, Jose-Maria Salinas, and Maria De~La
  Iglesia-Vaya.
\newblock Padchest: A large chest x-ray image dataset with multi-label
  annotated reports.
\newblock \emph{Medical image analysis}, 66:\penalty0 101797, 2020.

\bibitem[d'Eon et~al.(2022)d'Eon, d'Eon, Wright, and
  Leyton-Brown]{d2022spotlight}
Greg d'Eon, Jason d'Eon, James~R Wright, and Kevin Leyton-Brown.
\newblock The spotlight: A general method for discovering systematic errors in
  deep learning models.
\newblock In \emph{Proceedings of the 2022 ACM Conference on Fairness,
  Accountability, and Transparency}, pages 1962--1981, 2022.

\bibitem[Eyuboglu et~al.(2022)Eyuboglu, Varma, Saab, Delbrouck, Lee-Messer,
  Dunnmon, Zou, and R{\'e}]{eyuboglu2022domino}
Sabri Eyuboglu, Maya Varma, Khaled Saab, Jean-Benoit Delbrouck, Christopher
  Lee-Messer, Jared Dunnmon, James Zou, and Christopher R{\'e}.
\newblock Domino: Discovering systematic errors with cross-modal embeddings.
\newblock \emph{arXiv preprint arXiv:2203.14960}, 2022.

\bibitem[Gichoya et~al.(2023)Gichoya, Thomas, Celi, Safdar, Banerjee, Banja,
  Seyyed-Kalantari, Trivedi, and Purkayastha]{gichoya2023ai}
Judy~Wawira Gichoya, Kaesha Thomas, Leo~Anthony Celi, Nabile Safdar, Imon
  Banerjee, John~D Banja, Laleh Seyyed-Kalantari, Hari Trivedi, and Saptarshi
  Purkayastha.
\newblock Ai pitfalls and what not to do: mitigating bias in ai.
\newblock \emph{The British Journal of Radiology}, 96\penalty0 (1150):\penalty0
  20230023, 2023.

\bibitem[Guo et~al.(2018)Guo, Frank, and Weinberger]{guo2018low}
Chuan Guo, Jared~S Frank, and Kilian~Q Weinberger.
\newblock Low frequency adversarial perturbation.
\newblock \emph{arXiv preprint arXiv:1809.08758}, 2018.

\bibitem[He et~al.(2015)He, Zhang, Ren, and
  Sun]{he2015deepresiduallearningimage}
Kaiming He, Xiangyu Zhang, Shaoqing Ren, and Jian Sun.
\newblock Deep residual learning for image recognition, 2015.
\newblock URL \url{https://arxiv.org/abs/1512.03385}.

\bibitem[Hong et~al.(2024)Hong, Yue, Chen, Cong, Lin, Luo, Wang, Wang, Xu,
  Yang, et~al.]{hong2024out}
Zesheng Hong, Yubiao Yue, Yubin Chen, Lele Cong, Huanjie Lin, Yuanmei Luo,
  Mini~Han Wang, Weidong Wang, Jialong Xu, Xiaoqi Yang, et~al.
\newblock Out-of-distribution detection in medical image analysis: A survey.
\newblock \emph{arXiv preprint arXiv:2404.18279}, 2024.

\bibitem[Huang et~al.(2018)Huang, Liu, van~der Maaten, and
  Weinberger]{huang2018denselyconnectedconvolutionalnetworks}
Gao Huang, Zhuang Liu, Laurens van~der Maaten, and Kilian~Q. Weinberger.
\newblock Densely connected convolutional networks, 2018.
\newblock URL \url{https://arxiv.org/abs/1608.06993}.

\bibitem[Huang et~al.(2021{\natexlab{a}})Huang, Guan, Xiao, and
  Lu]{huang2021fsdr}
Jiaxing Huang, Dayan Guan, Aoran Xiao, and Shijian Lu.
\newblock Fsdr: Frequency space domain randomization for domain generalization.
\newblock In \emph{Proceedings of the IEEE/CVF conference on computer vision
  and pattern recognition}, pages 6891--6902, 2021{\natexlab{a}}.

\bibitem[Huang et~al.(2021{\natexlab{b}})Huang, Guan, Xiao, and
  Lu]{huang2021rda}
Jiaxing Huang, Dayan Guan, Aoran Xiao, and Shijian Lu.
\newblock Rda: Robust domain adaptation via fourier adversarial attacking.
\newblock In \emph{Proceedings of the IEEE/CVF international conference on
  computer vision}, pages 8988--8999, 2021{\natexlab{b}}.

\bibitem[Irvin et~al.(2019)Irvin, Rajpurkar, Ko, Yu, Ciurea-Ilcus, Chute,
  Marklund, Haghgoo, Ball, Shpanskaya, et~al.]{irvin2019chexpert}
Jeremy Irvin, Pranav Rajpurkar, Michael Ko, Yifan Yu, Silviana Ciurea-Ilcus,
  Chris Chute, Henrik Marklund, Behzad Haghgoo, Robyn Ball, Katie Shpanskaya,
  et~al.
\newblock Chexpert: A large chest radiograph dataset with uncertainty labels
  and expert comparison.
\newblock In \emph{Proceedings of the AAAI conference on artificial
  intelligence}, volume~33, pages 590--597, 2019.

\bibitem[Johnson et~al.(2019)Johnson, Pollard, Berkowitz, Greenbaum, Lungren,
  Deng, Mark, and Horng]{johnson2019mimic}
Alistair~EW Johnson, Tom~J Pollard, Seth~J Berkowitz, Nathaniel~R Greenbaum,
  Matthew~P Lungren, Chih-ying Deng, Roger~G Mark, and Steven Horng.
\newblock Mimic-cxr, a de-identified publicly available database of chest
  radiographs with free-text reports.
\newblock \emph{Scientific data}, 6\penalty0 (1):\penalty0 317, 2019.

\bibitem[Katzman et~al.(2024)Katzman, Alabousi, Islam, Zha, and
  Patlas]{katzman2024deep}
Benjamin~D Katzman, Mostafa Alabousi, Nabil Islam, Nanxi Zha, and Michael~N
  Patlas.
\newblock Deep learning for pneumothorax detection on chest radiograph: a
  diagnostic test accuracy systematic review and meta analysis.
\newblock \emph{Canadian Association of Radiologists Journal}, 75\penalty0
  (3):\penalty0 525--533, 2024.

\bibitem[Kingma and Ba(2017)]{kingma2017adammethodstochasticoptimization}
Diederik~P. Kingma and Jimmy Ba.
\newblock Adam: A method for stochastic optimization, 2017.
\newblock URL \url{https://arxiv.org/abs/1412.6980}.

\bibitem[Larrazabal et~al.(2020)Larrazabal, Nieto, Peterson, Milone, and
  Ferrante]{larrazabal2020gender}
Agostina~J Larrazabal, Nicol{\'a}s Nieto, Victoria Peterson, Diego~H Milone,
  and Enzo Ferrante.
\newblock Gender imbalance in medical imaging datasets produces biased
  classifiers for computer-aided diagnosis.
\newblock \emph{Proceedings of the National Academy of Sciences}, 117\penalty0
  (23):\penalty0 12592--12594, 2020.

\bibitem[Lee et~al.(2018)Lee, Lee, Lee, and Shin]{lee2018simple}
Kimin Lee, Kibok Lee, Honglak Lee, and Jinwoo Shin.
\newblock A simple unified framework for detecting out-of-distribution samples
  and adversarial attacks.
\newblock \emph{Advances in neural information processing systems}, 31, 2018.

\bibitem[Liu et~al.(2020)Liu, Wang, Owens, and Li]{liu2020energy}
Weitang Liu, Xiaoyun Wang, John Owens, and Yixuan Li.
\newblock Energy-based out-of-distribution detection.
\newblock \emph{Advances in neural information processing systems},
  33:\penalty0 21464--21475, 2020.

\bibitem[Oakden-Rayner et~al.(2020)Oakden-Rayner, Dunnmon, Carneiro, and
  R{\'e}]{oakden2020hidden}
Luke Oakden-Rayner, Jared Dunnmon, Gustavo Carneiro, and Christopher R{\'e}.
\newblock Hidden stratification causes clinically meaningful failures in
  machine learning for medical imaging.
\newblock In \emph{Proceedings of the ACM conference on health, inference, and
  learning}, pages 151--159, 2020.

\bibitem[Olesen et~al.(2024)Olesen, Weng, Feragen, and
  Petersen]{olesen2024slicing}
Vincent Olesen, Nina Weng, Aasa Feragen, and Eike Petersen.
\newblock Slicing through bias: explaining performance gaps in medical image
  analysis using slice discovery methods.
\newblock In \emph{MICCAI Workshop on Fairness of AI in Medical Imaging}, pages
  3--13. Springer, 2024.

\bibitem[Roy et~al.(2022)Roy, Ren, Azizi, Loh, Natarajan, Mustafa, Pawlowski,
  Freyberg, Liu, Beaver, et~al.]{roy2022does}
Abhijit~Guha Roy, Jie Ren, Shekoofeh Azizi, Aaron Loh, Vivek Natarajan, Basil
  Mustafa, Nick Pawlowski, Jan Freyberg, Yuan Liu, Zach Beaver, et~al.
\newblock Does your dermatology classifier know what it doesn’t know?
  detecting the long-tail of unseen conditions.
\newblock \emph{Medical Image Analysis}, 75:\penalty0 102274, 2022.

\bibitem[Seyyed-Kalantari et~al.(2020)Seyyed-Kalantari, Liu, McDermott, Chen,
  and Ghassemi]{seyyed2020chexclusion}
Laleh Seyyed-Kalantari, Guanxiong Liu, Matthew McDermott, Irene~Y Chen, and
  Marzyeh Ghassemi.
\newblock Chexclusion: Fairness gaps in deep chest x-ray classifiers.
\newblock In \emph{BIOCOMPUTING 2021: proceedings of the Pacific symposium},
  pages 232--243. World Scientific, 2020.

\bibitem[Tan and Le(2020)]{tan2020efficientnetrethinkingmodelscaling}
Mingxing Tan and Quoc~V. Le.
\newblock Efficientnet: Rethinking model scaling for convolutional neural
  networks, 2020.
\newblock URL \url{https://arxiv.org/abs/1905.11946}.

\bibitem[Tardy et~al.(2019)Tardy, Scheffer, and Mateus]{tardy2019uncertainty}
Mickael Tardy, Bruno Scheffer, and Diana Mateus.
\newblock Uncertainty measurements for the reliable classification of
  mammograms.
\newblock In \emph{International conference on medical image computing and
  computer-assisted intervention}, pages 495--503. Springer, 2019.

\bibitem[Torralba and Oliva(2003)]{torralba2003statistics}
Antonio Torralba and Aude Oliva.
\newblock Statistics of natural image categories.
\newblock \emph{Network: computation in neural systems}, 14\penalty0
  (3):\penalty0 391, 2003.

\bibitem[Wang et~al.(2020)Wang, Wu, Huang, and Xing]{wang2020high}
Haohan Wang, Xindi Wu, Zeyi Huang, and Eric~P Xing.
\newblock High-frequency component helps explain the generalization of
  convolutional neural networks.
\newblock In \emph{Proceedings of the IEEE/CVF conference on computer vision
  and pattern recognition}, pages 8684--8694, 2020.

\bibitem[Wang et~al.(2017)Wang, Peng, Lu, Lu, Bagheri, and
  Summers]{wang2017chestx}
Xiaosong Wang, Yifan Peng, Le~Lu, Zhiyong Lu, Mohammadhadi Bagheri, and
  Ronald~M Summers.
\newblock Chestx-ray8: Hospital-scale chest x-ray database and benchmarks on
  weakly-supervised classification and localization of common thorax diseases.
\newblock In \emph{Proceedings of the IEEE conference on computer vision and
  pattern recognition}, pages 2097--2106, 2017.

\bibitem[Wiles et~al.(2021)Wiles, Gowal, Stimberg, Alvise-Rebuffi, Ktena,
  Dvijotham, and Cemgil]{wiles2021fine}
Olivia Wiles, Sven Gowal, Florian Stimberg, Sylvestre Alvise-Rebuffi, Ira
  Ktena, Krishnamurthy Dvijotham, and Taylan Cemgil.
\newblock A fine-grained analysis on distribution shift.
\newblock \emph{arXiv preprint arXiv:2110.11328}, 2021.

\bibitem[Xu et~al.(2021)Xu, Zhang, Zhang, Wang, and Tian]{xu2021fourier}
Qinwei Xu, Ruipeng Zhang, Ya~Zhang, Yanfeng Wang, and Qi~Tian.
\newblock A fourier-based framework for domain generalization.
\newblock In \emph{Proceedings of the IEEE/CVF conference on computer vision
  and pattern recognition}, pages 14383--14392, 2021.

\bibitem[Yang et~al.(2022)Yang, Guo, Chen, and Yuan]{yang2022source}
Chen Yang, Xiaoqing Guo, Zhen Chen, and Yixuan Yuan.
\newblock Source free domain adaptation for medical image segmentation with
  fourier style mining.
\newblock \emph{Medical Image Analysis}, 79:\penalty0 102457, 2022.

\bibitem[Yu et~al.(2022)Yu, Mohajer, and Eng]{yu2022external}
Alice~C Yu, Bahram Mohajer, and John Eng.
\newblock External validation of deep learning algorithms for radiologic
  diagnosis: a systematic review.
\newblock \emph{Radiology: Artificial Intelligence}, 4\penalty0 (3):\penalty0
  e210064, 2022.

\bibitem[Zech et~al.(2018)Zech, Badgeley, Liu, Costa, Titano, and
  Oermann]{zech2018variable}
John~R Zech, Marcus~A Badgeley, Manway Liu, Anthony~B Costa, Joseph~J Titano,
  and Eric~Karl Oermann.
\newblock Variable generalization performance of a deep learning model to
  detect pneumonia in chest radiographs: a cross-sectional study.
\newblock \emph{PLoS medicine}, 15\penalty0 (11):\penalty0 e1002683, 2018.

\bibitem[Zhao et~al.(2022)Zhao, Liu, Sicilia, Hwang, and Fu]{zhao2022test}
Xingchen Zhao, Chang Liu, Anthony Sicilia, Seong~Jae Hwang, and Yun Fu.
\newblock Test-time fourier style calibration for domain generalization.
\newblock \emph{arXiv preprint arXiv:2205.06427}, 2022.

\end{thebibliography}

\clearpage

\appendix

\section{Ablation extension}\label{abla_ext}

\begin{figure}[!htbp]
\floatconts
  {fig:ablation_conso}
  {\caption{Ablation study evaluating the influence of the distance metric used in k-medoids clustering on AUROC variation for the Consolidation pathology. We compare three metrics: Euclidean ($L_2$) distance, Spearman distance, and the correlation distance employed in our proposed method. }}
  {
    \centering
    \begin{tabular}{ccc}
      \includegraphics[width=0.3\linewidth]{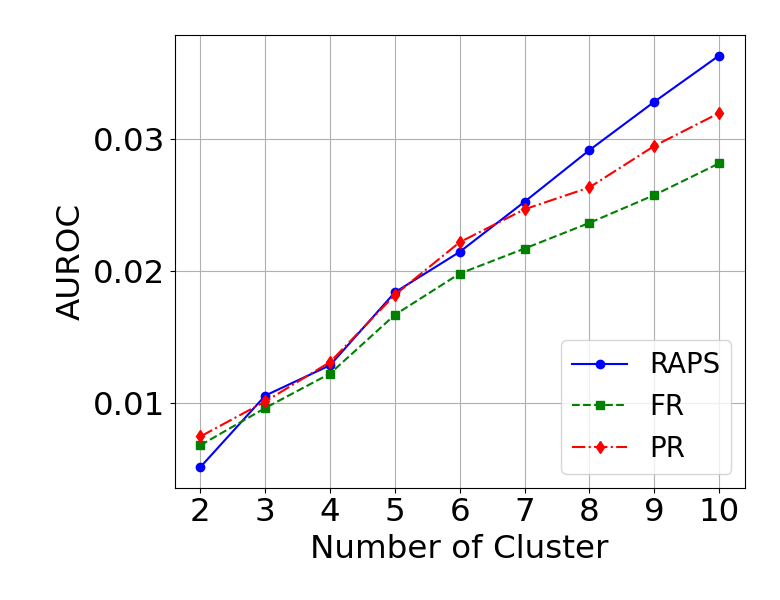} &

      \includegraphics[width=0.3\linewidth]{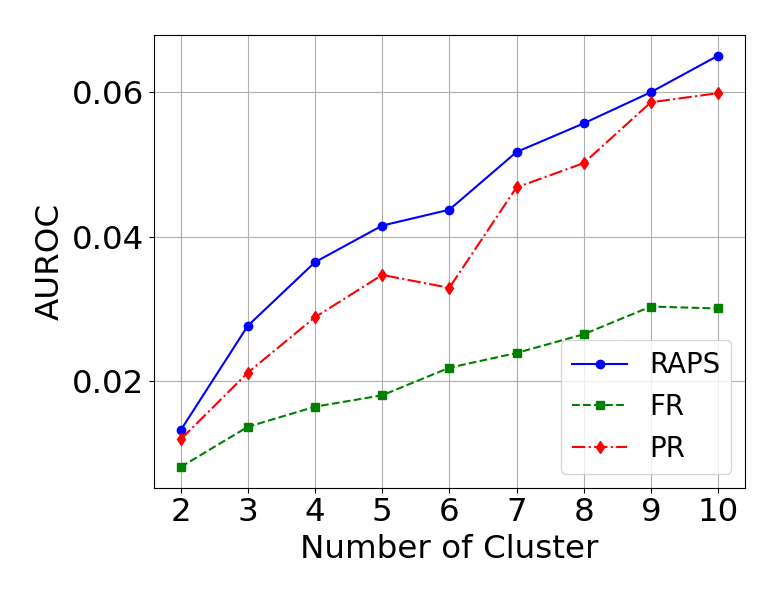} &

      \includegraphics[width=0.3\linewidth]{images/exp2/kmedoids/corr/kmedoids_plot_correlation__ONLYCHEXPERT_CORR_all_seed_std_stats_n50_filtred_train_consolidation_landscape_fromImagenet.png} \\
      
      (a) L2 & (b) Spearman & (c) Correlation
    \end{tabular}
  }
\end{figure}

\begin{figure}[!htbp]
\floatconts
  {fig:ablation_pleural}
  {\caption{Ablation study evaluating the influence of the distance metric used in k-medoids clustering on AUROC variation for the Pleural Effusion pathology. We compare three metrics: Euclidean ($L_2$) distance, Spearman distance, and the correlation distance employed in our proposed method.}}
  {%
    \centering
    \begin{tabular}{ccc}
      \includegraphics[width=0.3\linewidth]{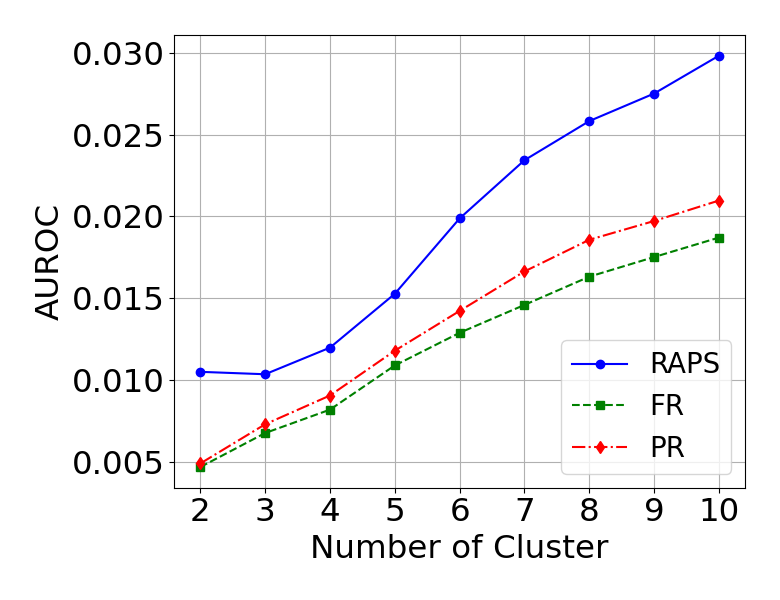} &

      \includegraphics[width=0.3\linewidth]{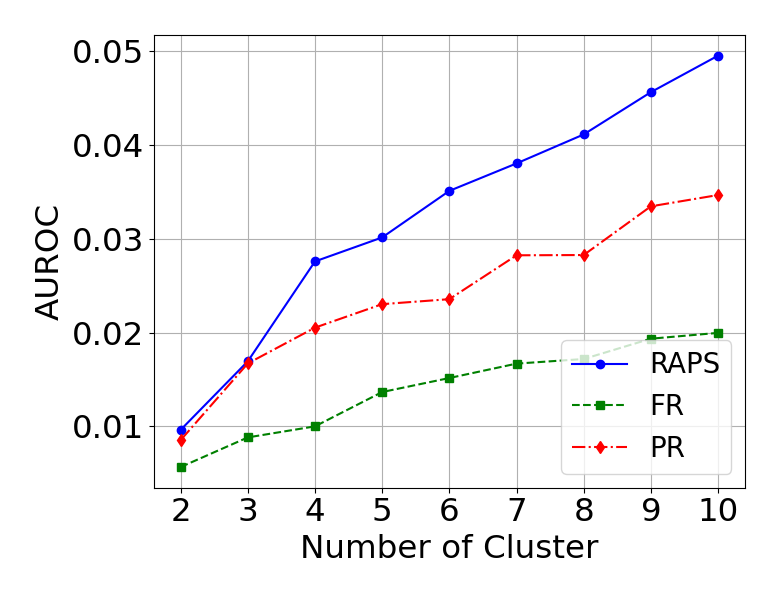} &

      \includegraphics[width=0.3\linewidth]{images/exp2/kmedoids/corr/kmedoids_plot_correlation__ONLYCHEXPERT_CORR_all_seed_std_stats_n50_filtred_train_pleuralEffusion_landscape_fromImagenet.png} \\
      
      (a) L2 & (b) Spearman & (c) Correlation
    \end{tabular}
  }
\end{figure}\textbf{}

\clearpage

\section{Results with other models}\label{more_models}

\begin{figure}[!htbp]
\floatconts
  {fig:kmedoids2_corr_resnet}
  {\caption{RAPS evaluated on Resnet50 model \cite{he2015deepresiduallearningimage}. AUROC variability across different number of clusters ($K$) made by the k-medoids method with the correlation metric vs. clusters made by fully and pseudo random methods. Results are averaged over 50 seeds.}}
  {%
    \centering
    \begin{tabular}{ccc}
      \includegraphics[width=0.3\linewidth]{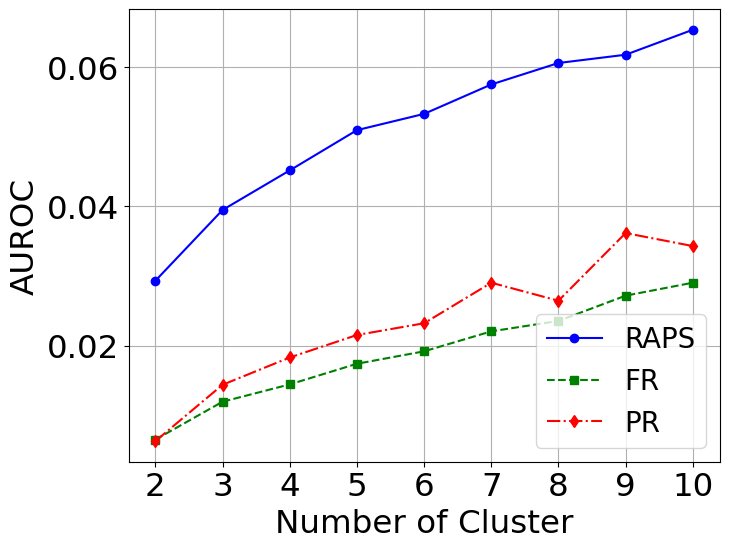} &
      \includegraphics[width=0.3\linewidth]{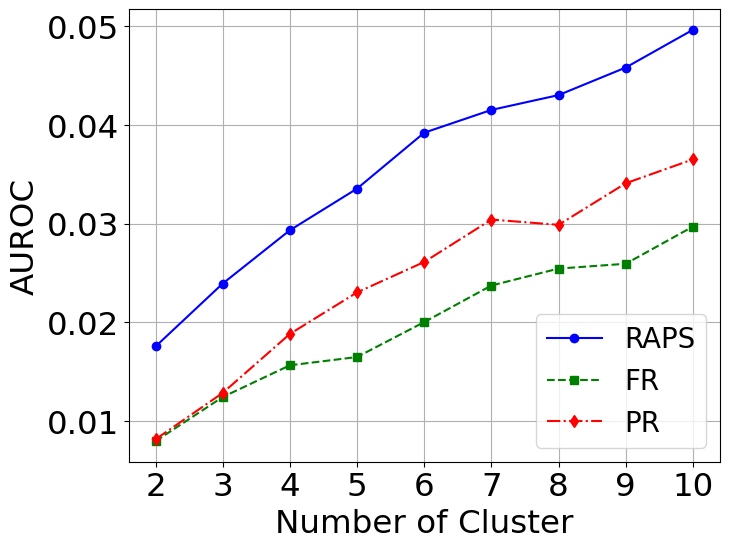} &
      \includegraphics[width=0.3\linewidth]{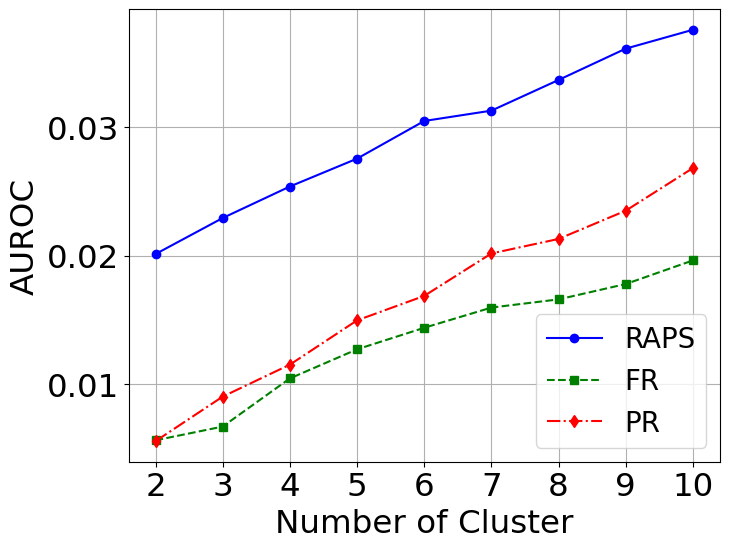} \\
      (a) Atelectasis & (b) Consolidation & (c) Pleural Effusion
    \end{tabular}
  }
\end{figure}

\begin{figure}[!htbp]
\floatconts
  {fig:kmedoids2_corr_efficientnet}
  {\caption{RAPS evaluated on EfficientNet model \cite{tan2020efficientnetrethinkingmodelscaling}. AUROC variability across different number of clusters ($K$) made by the k-medoids method with the correlation metric vs. clusters made by fully and pseudo random methods. Results are averaged over 50 seeds.}}
  {%
    \centering
    \begin{tabular}{ccc}
      \includegraphics[width=0.3\linewidth]{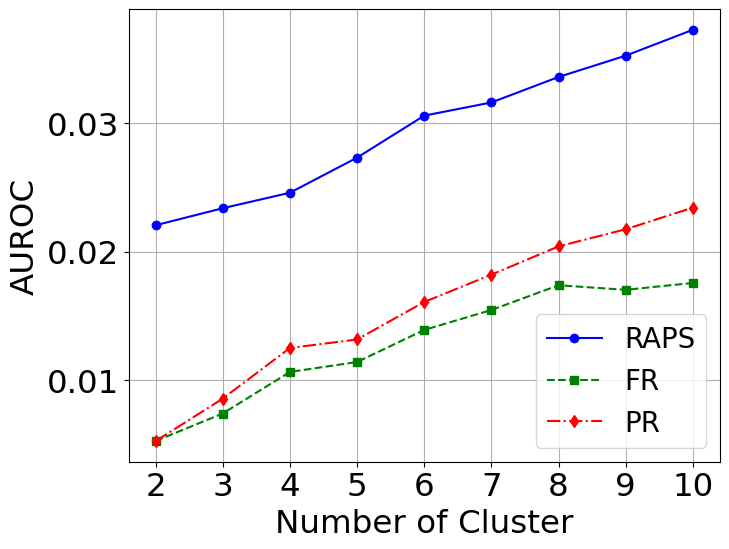} &
      \includegraphics[width=0.3\linewidth]{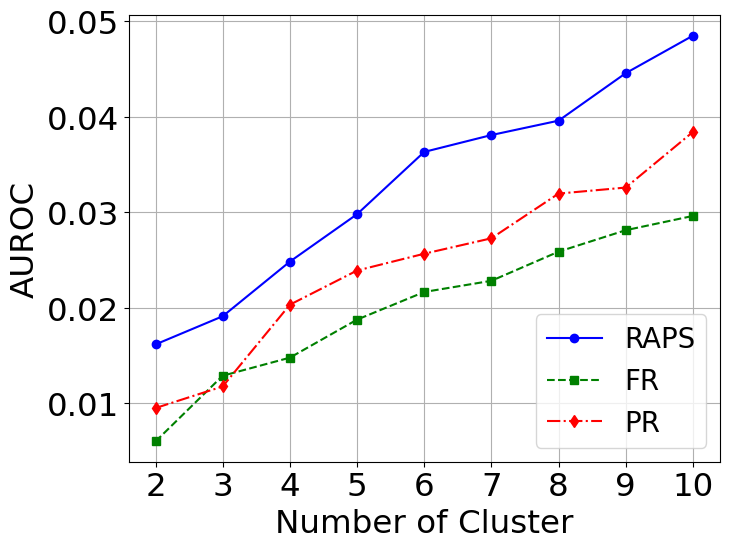} &
      \includegraphics[width=0.3\linewidth]{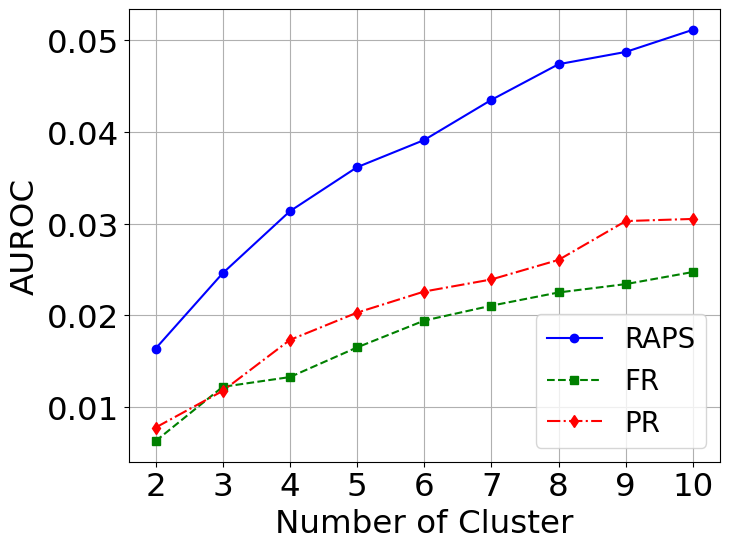} \\
      (a) Atelectasis & (b) Consolidation & (c) Pleural Effusion
    \end{tabular}
  }
\end{figure}

\newpage
\section{Comparison with semantic metadata}\label{slice_comparison_apendix}

\color{black}
For Pleural Effusion cases from the NIH center, we compare the AUROC of the worst slice isolated by our method with traditional slicing by available metadata attributes, i.e. age, gender, and view. As demonstrated in Table \ref{tab:slice_comparison}, metadata slicing sometimes fails to identify performance-degrading subsets, resulting in only minor AUROC variations.
In contrast, our data-centric framework successfully isolates a highly vulnerable data slice that exhibits a substantial and critical performance drop.
This result definitively establishes our method as an effective complementary solution for automated failure mode discovery.

\begin{table}[!htbp]
\floatconts
  {tab:slice_comparison}
  {\caption{Comparison of model performance achieved by slicing the data using conventional semantic metadata against the proposed HAC of RAPS method. Results are presented for the Pleural Effusion pathology within the NIH dataset. (P: Posterior; A: Anterior)}}
  {\scalebox{0.8}{\begin{tabular}{llc}
    \hline
    \textbf{Category} & \textbf{Subgroup} & \textbf{AUROC} \\
    \hline
    Age & $<20$  & 0.863 \\
        & $20\text{--}60$  & 0.832 \\
        & $>60$  & 0.839 \\
    \hline
    Gender & Female  & 0.856 \\
           & Male    & 0.824 \\
    \hline
    View & AP  & 0.827 \\
         & PA  & 0.845 \\
    \hline
    \textbf{Proposed} & \textbf{Our Slice} & \textbf{0.645} \\
    \hline
    \end{tabular}}}
\end{table}

\newpage
\section{Qualitative analysis}\label{quali}

\begin{figure}[htbp]
\floatconts
  {fig:qualivis_atelectasis_pad}
  {\caption{Qualitative results on Atelectasis for Padchest.}}
  {\includegraphics[width=\linewidth]{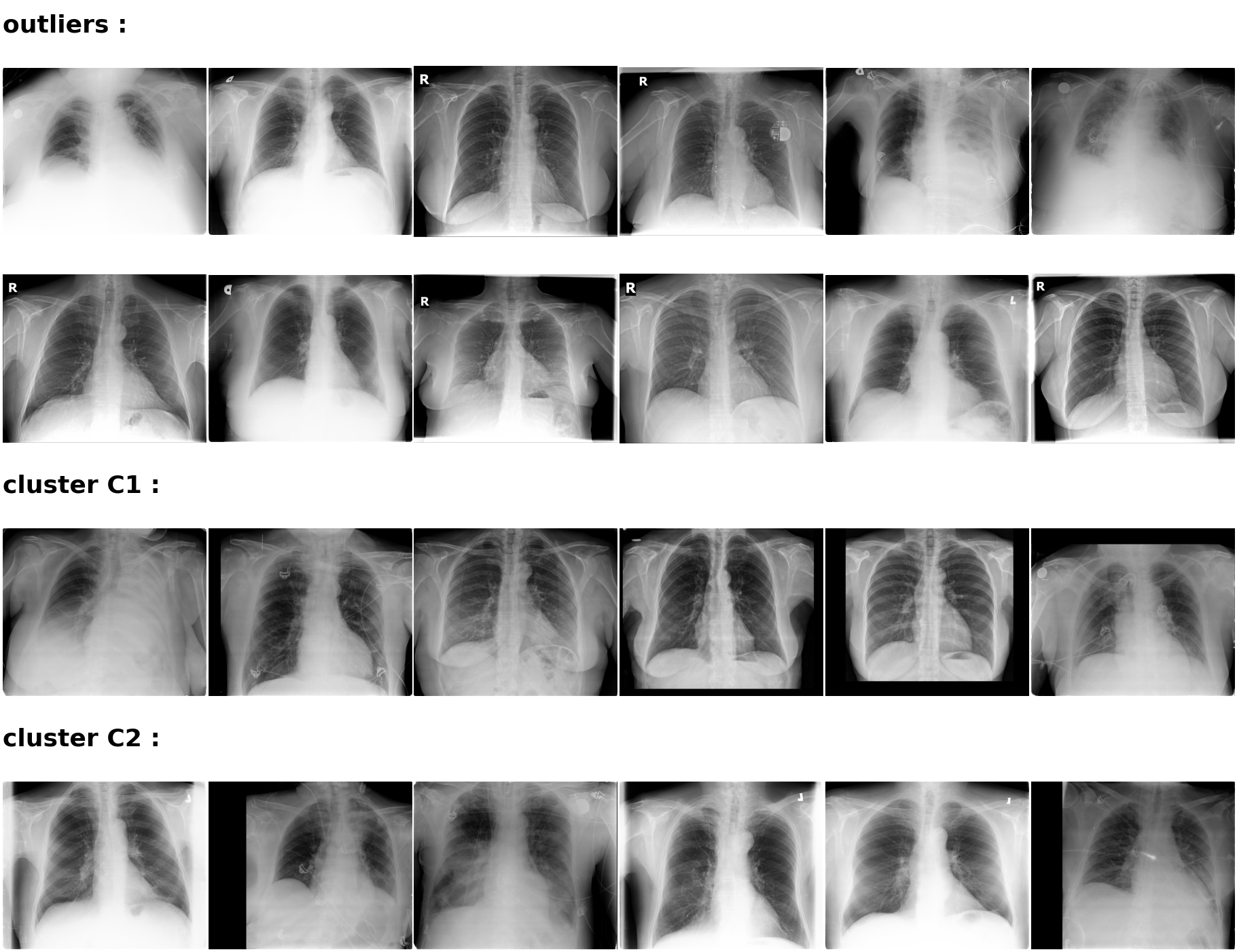}}
\end{figure}

\begin{figure}[htbp]
\floatconts
  {fig:qualivis_atelectasis_mimic}
  {\caption{Qualitative results on Atelectasis for MIMIC.}}
  {\includegraphics[width=\linewidth]{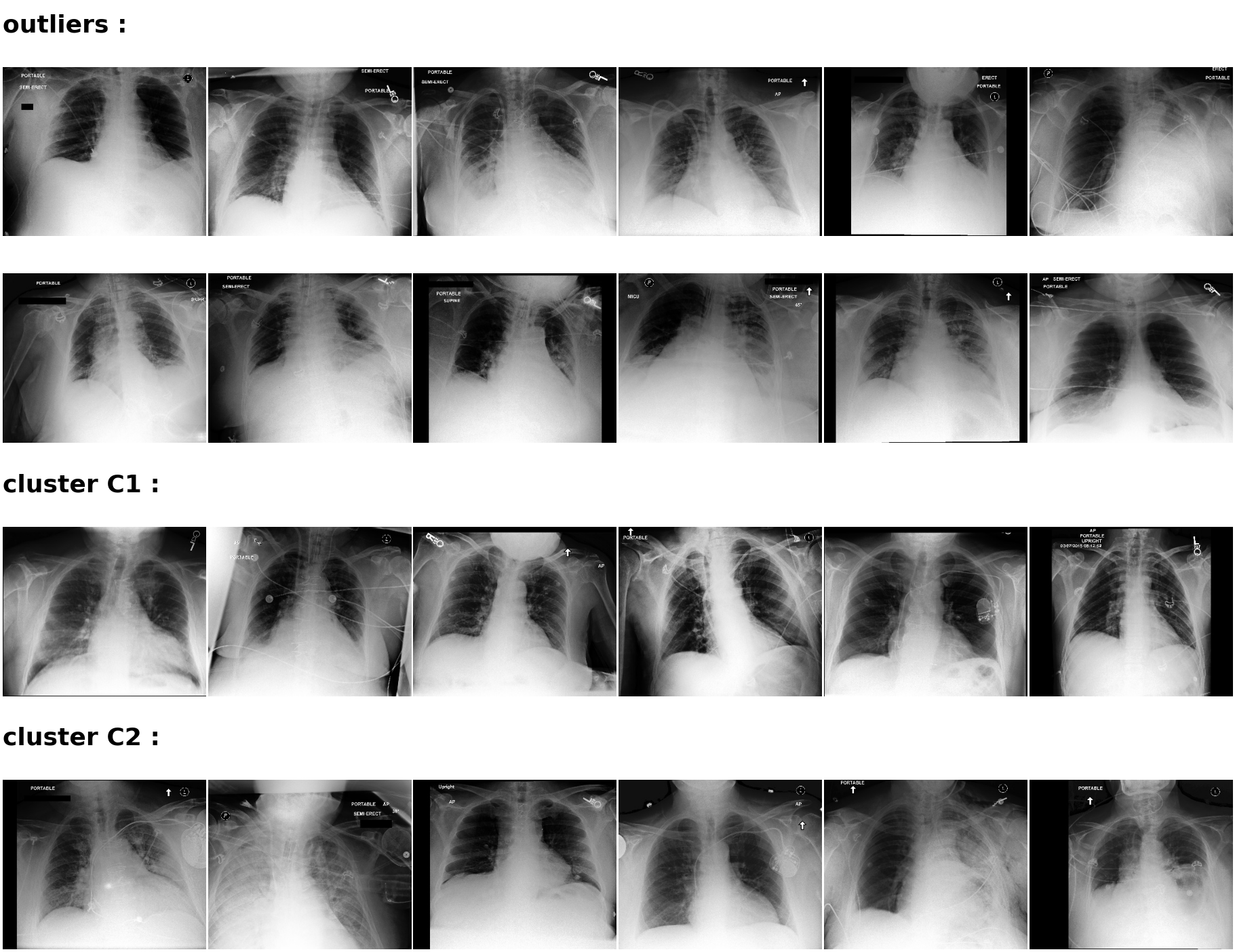}}
\end{figure}

\begin{figure}[htbp]
\floatconts
  {fig:qualivis_atelectasis_nih}
  {\caption{Qualitative results on Atelectasis for NIH.}}
  {\includegraphics[width=\linewidth]{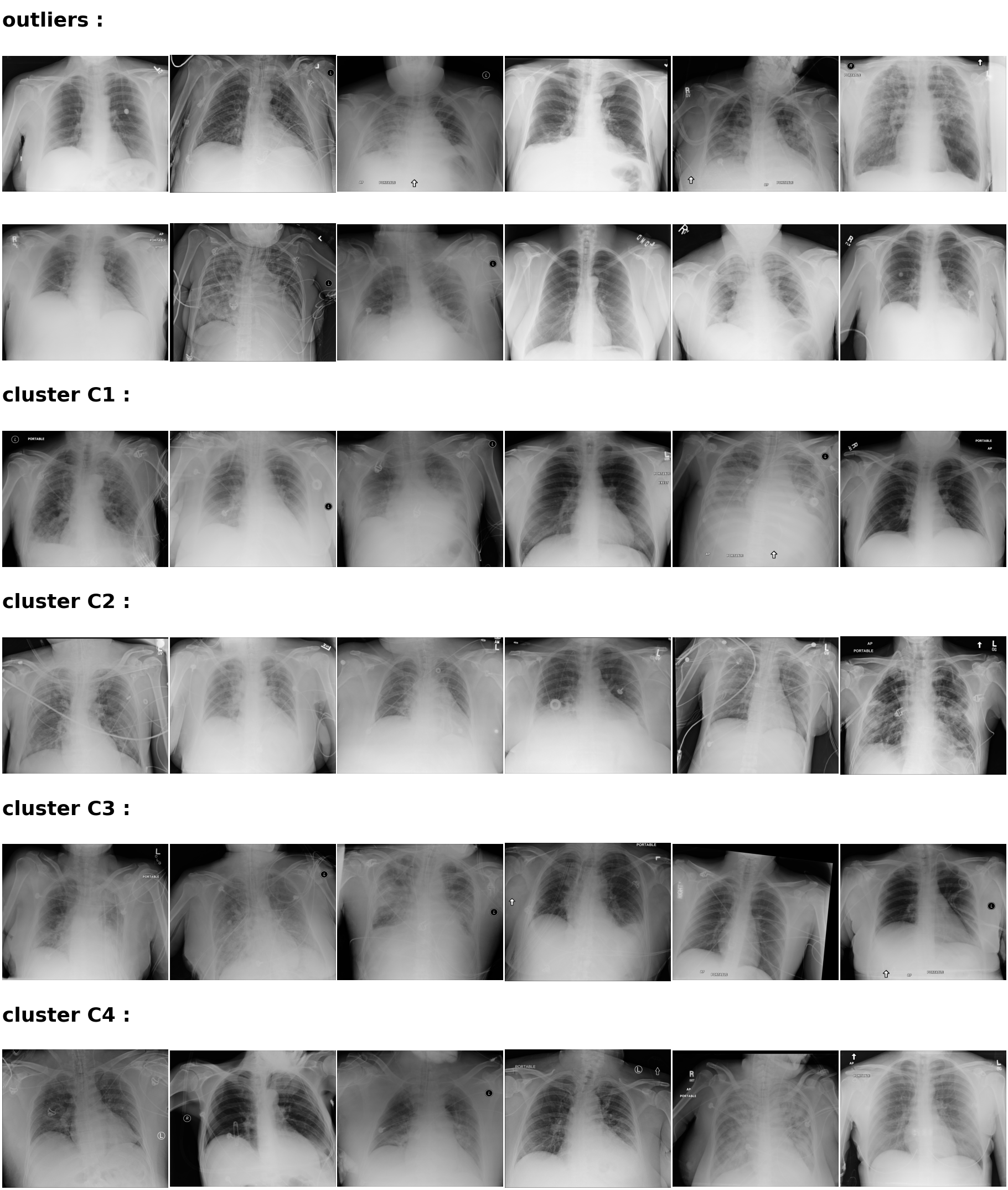}}
\end{figure}

\begin{figure}[htbp]
\floatconts
  {fig:qualivis_conso_pad}
  {\caption{Qualitative results on Consolidation for Padchest.}}
  {\includegraphics[width=\linewidth]{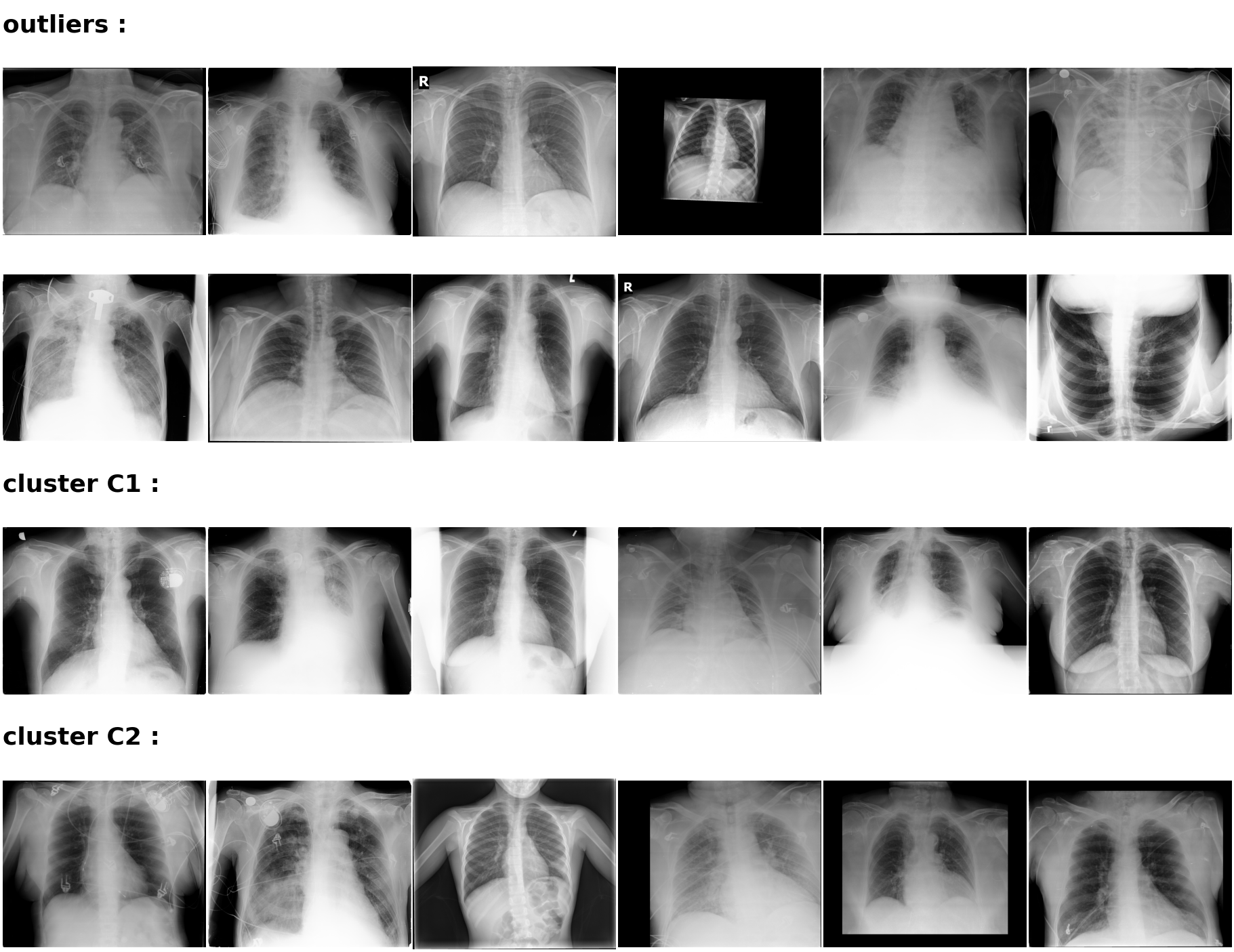}}
\end{figure}

\begin{figure}[htbp]
\floatconts
  {fig:qualivis_conso_mimic}
  {\caption{Qualitative results on Consolidation for MIMIC.}}
  {\includegraphics[width=\linewidth]{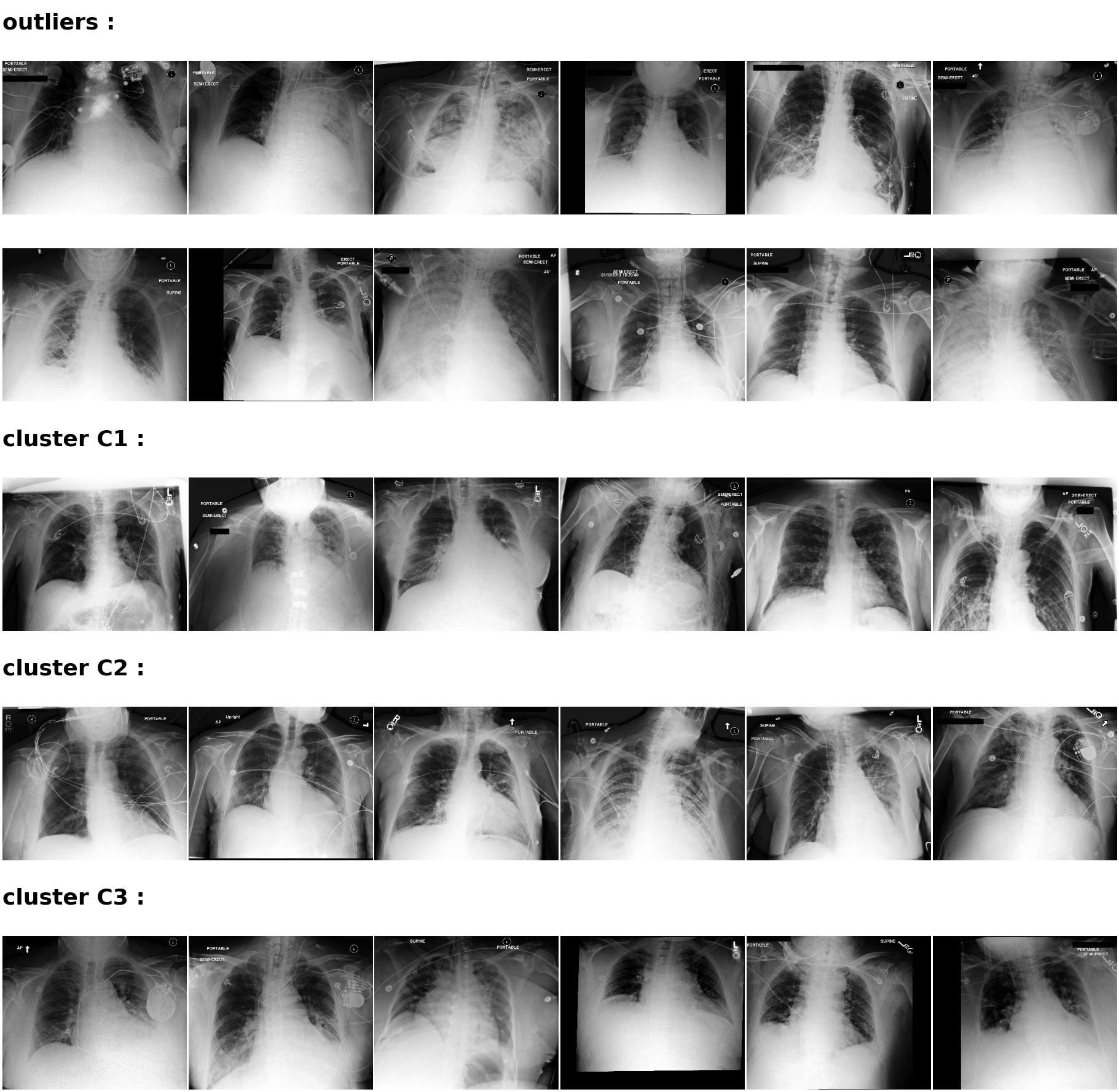}}
\end{figure}

\begin{figure}[!htbp]
\floatconts
  {fig:qualivis_conso_nih}
  {\caption{Qualitative results on Consolidation for NIH.}}
  {\includegraphics[width=\linewidth]{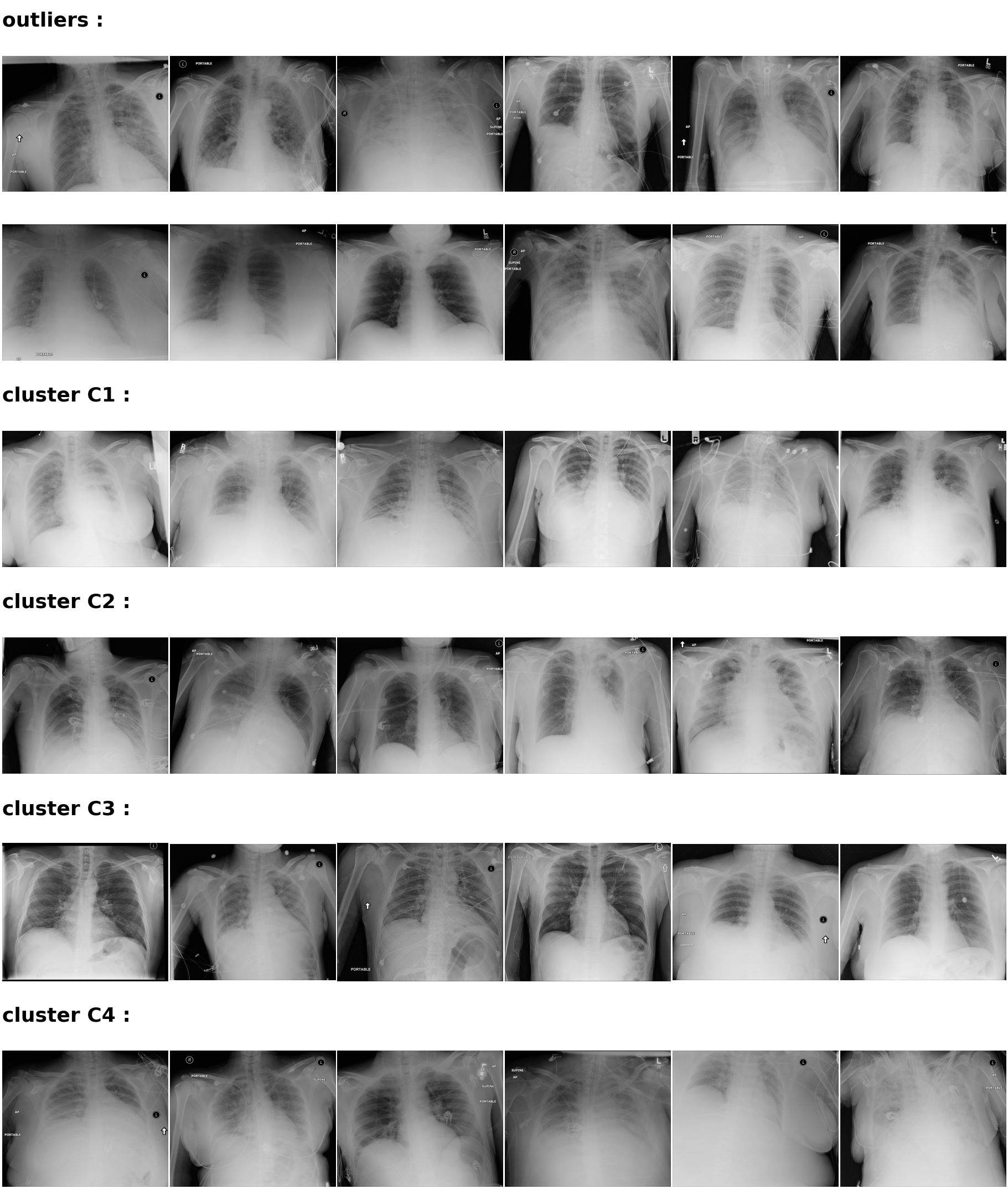}}
\end{figure}

\begin{figure}[!htbp]
\floatconts
  {fig:qualivis_pleural_pad}
  {\caption{Qualitative results on Pleural Effusion for Padchest.}}
  {\includegraphics[width=\linewidth]{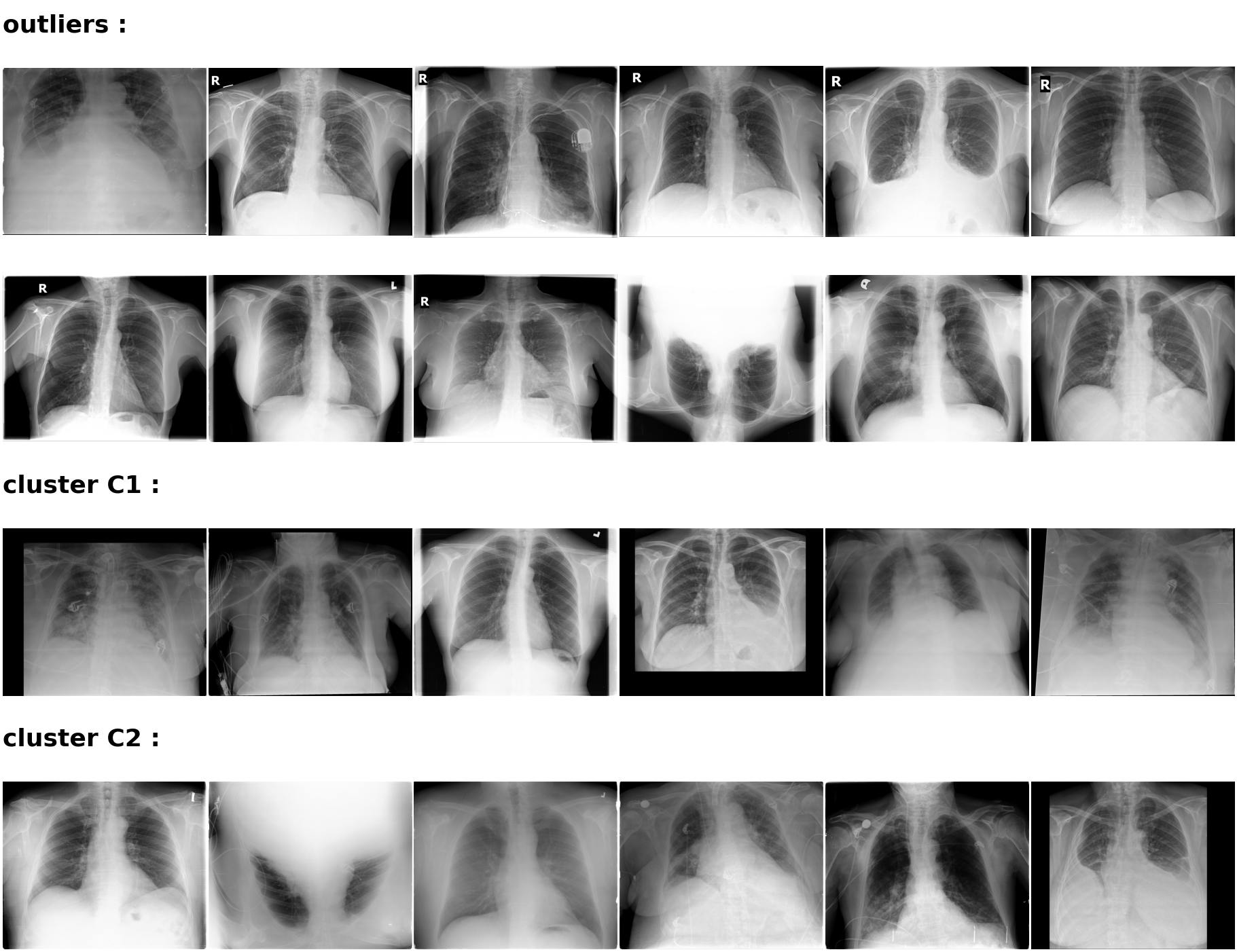}}
\end{figure}

\begin{figure}[!htbp]
\floatconts
  {fig:qualivis_pleural_mimic}
  {\caption{Qualitative results on Pleural Effusion for MIMIC.}}
  {\includegraphics[width=\linewidth]{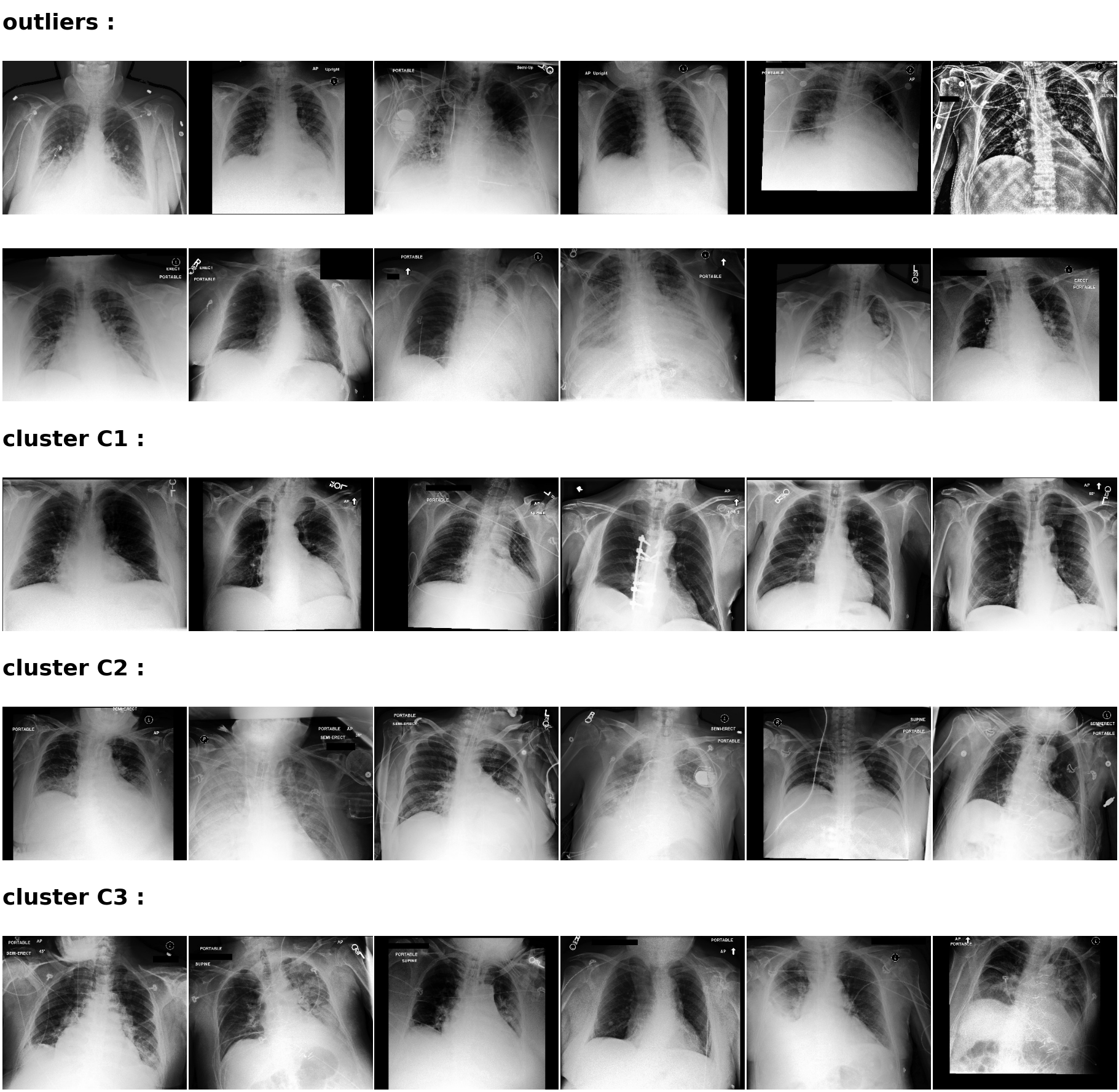}}
\end{figure}

\begin{figure}[!htbp]
\floatconts
  {fig:qualivis_pleural_nih}
  {\caption{Qualitative results on Pleural Effusion for NIH.}}
  {\includegraphics[width=0.74\linewidth]{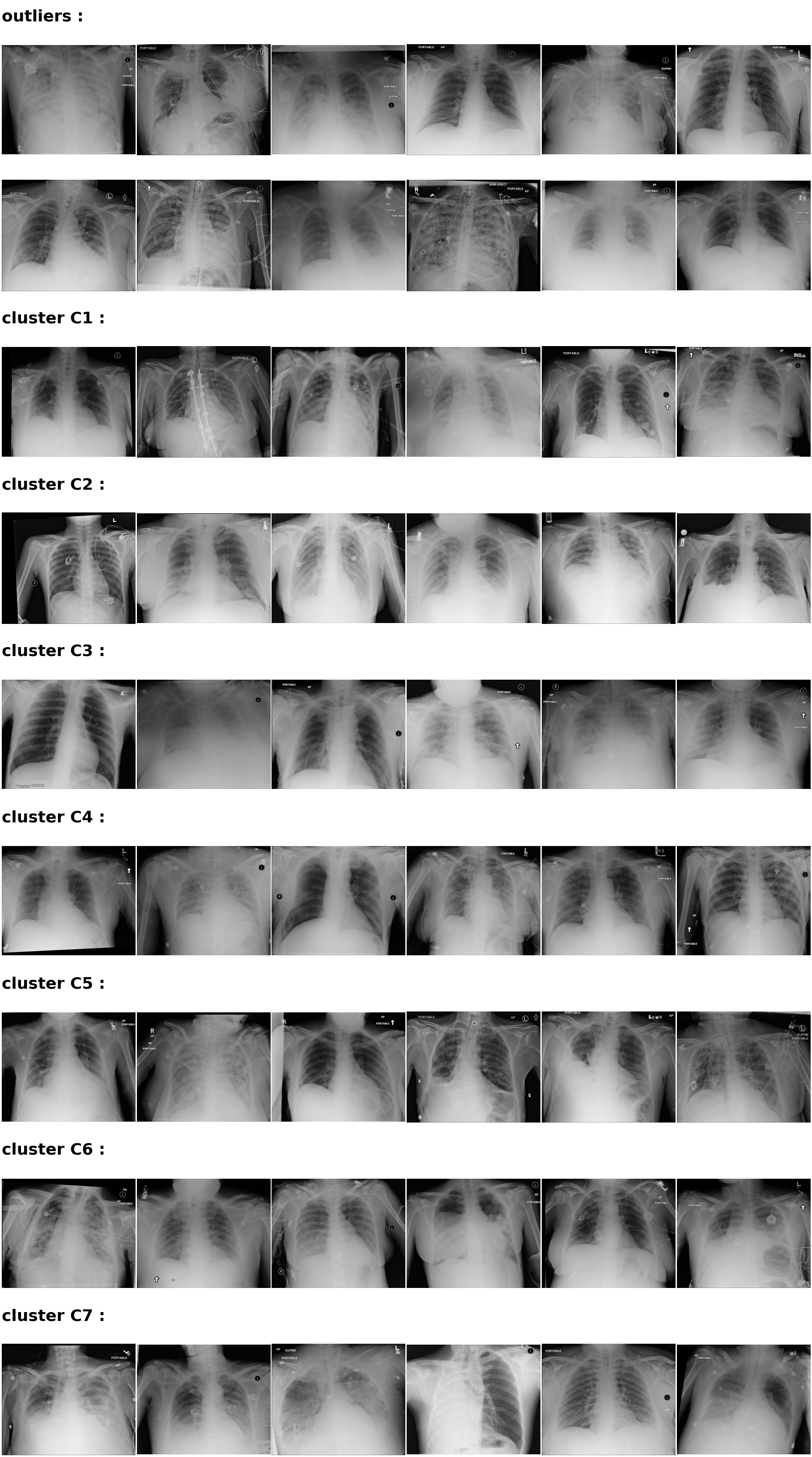}}
\end{figure}\end{document}